\documentclass[preprint,12pt]{elsarticle}

\usepackage{amssymb}
\usepackage{amsmath}
\usepackage{url}
\usepackage{textcomp}
\usepackage{booktabs}
\usepackage{xcolor}
\journal{Computers in Biology and Medicine}

\begin{document}

\begin{frontmatter}

%% Title, authors and addresses

%% use the tnoteref command within \title for footnotes;
%% use the tnotetext command for theassociated footnote;
%% use the fnref command within \author or \affiliation for footnotes;
%% use the fntext command for theassociated footnote;
%% use the corref command within \author for corresponding author footnotes;
%% use the cortext command for theassociated footnote;
%% use the ead command for the email address,
%% and the form \ead[url] for the home page:
%% \title{Title\tnoteref{label1}}
%% \tnotetext[label1]{}
%% \author{Name\corref{cor1}\fnref{label2}}
%% \ead{email address}
%% \ead[url]{home page}
%% \fntext[label2]{}
%% \cortext[cor1]{}
%% \affiliation{organization={},
%%             addressline={},
%%             city={},
%%             postcode={},
%%             state={},
%%             country={}}
%% \fntext[label3]{}

% \title{Machine Learning for Predicting Surgical Recommendations in Chronic Rhinosinusitis from Clinical Records}

\title{Machine Learning Based Prediction of Surgical Outcomes in Chronic Rhinosinusitis from Clinical Data}

%% use optional labels to link authors explicitly to addresses:
%% \author[label1,label2]{}
%% \affiliation[label1]{organization={},
%%             addressline={},
%%             city={},
%%             postcode={},
%%             state={},
%%             country={}}
%%
%% \affiliation[label2]{organization={},
%%             addressline={},
%%             city={},
%%             postcode={},
%%             state={},
%%             country={}}

\author[1,3]{Sayeed Shafayet Chowdhury}
\author[2]{Karen D'Souza}
\author[1]{V. Siva Kakumani}
\author[1]{Snehasis Mukhopadhyay}
\author[3]{Shiaofen Fang}
\author[4]{Rodney J. Schlosser}
\author[5]{Daniel M. Beswick}
\author[6]{Jeremiah A. Alt}
\author[7]{Jess C. Mace}
\author[4]{Zachary M. Soler}
\author[7]{Timothy L. Smith}
\author[9]{Vijay R. Ramakrishnan}

\affiliation[1]{organization={Purdue University}, city={Indianapolis}, state={IN}, country={USA}}
\affiliation[2]{organization={Idaho National Laboratory}, city={Idaho Falls}, state={ID}, country={USA}}
\affiliation[3]{organization={Indiana University Indianapolis}, city={Indianapolis}, state={IN}, country={USA}}
\affiliation[4]{organization={Department of Otolaryngology-Head and Neck Surgery, Medical University of South Carolina}, city={Charleston}, state={SC}, country={USA}}
\affiliation[5]{organization={Department of Otolaryngology-Head and Neck Surgery, University of California, Los Angeles}, city={Los Angeles}, state={CA}, country={USA}}
\affiliation[6]{organization={Department of Otolaryngology-Head and Neck Surgery, University of Utah}, city={Salt Lake City}, state={UT}, country={USA}}
\affiliation[7]{organization={Department of Otolaryngology-Head and Neck Surgery, Oregon Health Sciences University}, city={Portland}, state={OR}, country={USA}}
\affiliation[9]{organization={Department of Otolaryngology-Head and Neck Surgery, Indiana University School of Medicine}, city={Indianapolis}, state={IN}, country={USA}}

%% Abstract
\begin{abstract}
%% Text of abstract
Artificial intelligence (AI) has increasingly transformed medical data prognostics by enabling rapid and accurate analysis across imaging and pathology. However, investigation of machine learning predictions applied to prospectively collected, standardized, data from observational clinical intervention trials remains underexplored, despite its potential to reduce costs and improve patient outcomes. Chronic Rhinosinusitis (CRS), a persistent inflammatory disease of the paranasal sinuses that lasts more than three months, imposes a substantial burden on quality of life (QoL) and cost to society. Although many patients respond to medical therapy, others with refractory symptoms often pursue surgical intervention. Surgical decision-making in CRS is complex, as it must weigh known procedural risks against unclear individualized outcomes. In this study, we evaluated the utility of supervised machine learning models for predicting surgical benefit in CRS, using the main patient-reported outcome (Sino-Nasal Outcome Test-22, SNOT-22) as the primary outcome. Our cohort of prospectively collected data from an observational intervention trial comprised patients who all underwent surgery; we specifically investigate whether models trained only on preoperative data could have identified those who might not have been recommended surgery prior to the procedure. Across multiple algorithms, including an ensemble approach, our best model achieved $\sim$85\% classification accuracy, delivering accurate and interpretable predictions of surgical candidacy. Moreover, on a heldout set of 30 cases spanning mixed difficulty, our model achieved 80\% accuracy, exceeding the average prediction accuracy of expert clinicians (75.6\%), further demonstrating its potential to augment clinical decision-making and support personalized care in CRS management.

\end{abstract}

%%Graphical abstract
% \begin{graphicalabstract}
% %\includegraphics{grabs}
% \end{graphicalabstract}

% %%Research highlights
% \begin{highlights}
% \item Research highlight 1
% \item Research highlight 2
% \end{highlights}

%% Keywords
\begin{keyword}
%% keywords here, in the form: keyword \sep keyword
Chronic Rhinosinusitis \sep sinusitis \sep sinus surgery \sep ESS \sep SNOT22 \sep Surgical Outcomes \sep Supervised
Machine Learning \sep Binary Classification \sep Accuracy.
%% PACS codes here, in the form: \PACS code \sep code

%% MSC codes here, in the form: \MSC code \sep code
%% or \MSC[2008] code \sep code (2000 is the default)

\end{keyword}

\end{frontmatter}

%% Add \usepackage{lineno} before \begin{document} and uncomment 
%% following line to enable line numbers
%% \linenumbers

%% main text
%%

\section{Introduction}
\label{sec1:intro}

Artificial Intelligence (AI) for healthcare is emerging as one of the most dynamic areas of research and development worldwide. The foundational ideas date to the 1950s, when the Turing test was proposed as an operational benchmark for machine intelligence. Subsequent methodological and hardware advances over the years have enabled translational applications across health sciences \cite{Malik2019}. Today, AI and machine learning (ML) power a broad spectrum of products and services: from business analytics and robotics to voice-interactive assistants such as Siri, Alexa, and Google Assistant \cite{Mintz2019}. Healthcare, in particular, has seen rapid growth in AI investment and adoption since 2016, reflecting the potential of technology in this industry  to improve outcomes and reduce costs \cite{Malik2019,Buch2018,sahni2023artificial,varghese2024artificial,yu2018artificial}.

There are many uses for ML/AI in medical applications for clinical cases beyond basic operations. For clinicians, high-quality treatment recommendations depend on deep domain expertise and experience in interpreting complex heterogeneous information. As electronic health records (EHRs) have become the primary medium of documentation, clinicians spend substantial time on data entry, yet the resulting information, collected in part to enable billing and administrative workflows and in part to support clinical care and longitudinal analysis, remains difficult to synthesize during brief patient encounters. In addition, EHR data span both structured elements (e.g., demographics, diagnoses, medications, laboratory values, and standardized patient-reported outcomes) and unstructured elements (e.g., free-text clinical notes), each offering complementary signals whose utility depends on the study context and modeling objective. In this work, we focus on structured preoperative clinical variables and standardized patient-reported outcomes, which are readily comparable across patients and amenable to reproducible model development. For treatment recommendations, risk stratification, and outcome forecasting in personalized medicine, ML models can help shoulder this cognitive load by learning patterns from large datasets and producing fast, reproducible risk estimates or recommendations \cite{Mintz2019}. In one possible collaborative paradigm, AI augments, not replaces, clinical judgment: physicians and patients focus their limited time on interpreting model-informed options, clarifying trade-offs, and engaging in shared decision-making. Although AI has transformed several diagnostic domains (e.g., imaging and digital pathology) \cite{Suh2022}, the next sequential phase of use in the treatment domain remains comparatively underexplored.

Introducing AI-based data analytics to inform treatment selection enables physicians to synthesize data and findings learned from large, heterogeneous clinical datasets rapidly at the point-of-care. Objective, data-driven modeling can strengthen patient trust, help avoid unnecessary procedures, and support high-value, cost-effective care \cite{collins2015new,hellings2017positioning,muraro2016precision}. Neither AI nor the most experienced clinician can achieve $100\%$ accuracy; however, when used collaboratively, ML predictions can complement human expertise and improve decision quality, %augmenting rather than replacing clinicians,     
standardizing subjective data interpretation, potentially elevating less experienced non-specialty physicians \cite{Buch2018,Loftus2020JAMASurg}. In this study, we report the results of supervised ML modeling in one of the most prevalent and costly health problems in the United States, Chronic Rhinosinusitis (CRS), which affects a substantial fraction of adults and is a leading driver of outpatient visits, medication use, and overall healthcare utilization \cite{Rudmik2017,Orlandi2021ICAR}. When symptoms remain refractory to appropriate medical therapy, endoscopic sinus surgery is commonly pursued with the expectation of meaningful improvement in sinonasal quality of life; however, responses are heterogeneous and a non-trivial subset of patients experience limited benefit despite procedural risk and cost.
We evaluate supervised ML prediction models for supporting \emph{surgical recommendation} in CRS using SNOT-22 as the reference outcome. Importantly, our prospectively collected cohort comprises patients who all \emph{underwent surgery}; we specifically ask whether models trained solely on \emph{pre-operative} data could have \emph{preemptively} identified those who might not have been recommended surgery (limited SNOT-22 benefit = class 0), despite ultimately receiving it. We formulate this as a counterfactual classification task in which class 1 indicates ``surgery recommended as per the outcome of the prediction model'' and class 0 indicates ``surgery not recommended as per the outcome of the prediction model,'' with labels derived from 6-month SNOT-22 change relative to a minimal clinically important difference threshold \cite{Le2018,Soler2018Laryngoscope,Hopkins2018CHROME}.

We benchmark a suite of models, including logistic regression, support vector machines, random forests, Naïve Bayes, and neural networks, as well as ensemble strategies. Among these, a compact \emph{three-layer DNN (one hidden layer)} and a \emph{majority-voting ensemble} performed best, achieving \(\sim\)85\% classification accuracy. The main contributions of this work are:

\begin{itemize}%[leftmargin=0pt]
\setlength{\itemindent}{0em}

\item We frame CRS surgical selection as a preoperative surgical benefit prediction task, estimating the likelihood of achieving clinically meaningful improvement following ESS, training solely on \emph{pre-operative} variables.

\item We implement and compare multiple ML models and ensembles, demonstrating that AI is feasible for CRS surgery recommendation. To our knowledge, this is the first large-scale, multi-center study of its kind.

\item We conduct experiments to benchmark the performance of AI based algorithms to those of 6 human experts. Our results show that the proposed approach outperforms the experts on average in CRS surgical outcome prediction.

\end{itemize}

% Finally, we position the models as an adjunct to clinician judgment and shared decision-making, emphasizing calibrated probabilities, class-0 (do-not-operate) recall, and transparent explanations rather than deterministic rules \cite{Loftus2020JAMASurg,AAOHNSF2025Shin}.

\begin{figure}[t]
\centering
\includegraphics[width=0.5\columnwidth]{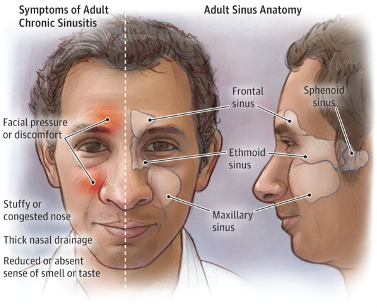}
\caption{Classic local manifestations of Chronic Rhinosinusitis (CRS). In addition, systemic such as fatigue and depression, and financial burden of direct and indirect costs, together result in a large overall disease burden (Figure reused with permission from \cite{rudmik2015medical}).}
\label{fig1}
\end{figure}

\section{Chronic Rhinosinusitis}

\subsection{Economics on CRS}
Chronic rhinosinusitis (CRS) is a persistent inflammatory disorder of the nose and paranasal sinuses, defined clinically by a constellation of local symptoms lasting $\geq 3$ months (Fig.~1) \cite{Orlandi2021ICAR}. In Western adult populations, its prevalence is commonly estimated at roughly 10--15\% \cite{Orlandi2021ICAR}. Although acute infections may precipitate or accompany disease flares, the hallmarks of CRS are \emph{chronic mucosal inflammation} with tissue remodeling and associated disturbances of mucus clearance and epithelial function \cite{stevens2015chronic}. Symptom burden is substantial: patients frequently report facial pain/pressure and headache, nasal obstruction and congestion, rhinorrhea/postnasal drainage, and loss of smell/taste, while systemic or non-nasal manifestations (e.g., ear pain, dental pain, cough, sleep disruption, fatigue, and mood symptoms) further compound individual quality-of-life (QoL) impairment. Beyond sinonasal symptoms, CRS is associated with broad health-related quality-of-life declines, exacerbation of comorbid atopic and respiratory diseases, work and school through absenteeism and presenteeism, and commonly involves repeated healthcare encounters and courses of antibiotics and systemic corticosteroids with attendant adverse effects and costs \cite{Rudmik2017,Orlandi2021ICAR,Choi2024Comorbidities,Rudmik2014Productivity,Pandrangi2023WPAI,Smith2022Antibiotics,Davis2022SCS}. Overall QoL decrements in CRS are comparable to those seen with major chronic conditions such as congestive heart failure, chronic obstructive pulmonary disease, end-stage renal disease on dialysis, and chronic low back pain \cite{Soler2011HSUV}. The societal impact is likewise considerable, with costs accruing from direct medical care and productivity losses; per-patient annual economic burden in the United States has been estimated at approximately \$10{,}077 \cite{Rudmik2017}.

\subsection{CRS Disease Management}
Initial treatment consists of medical treatments, including topical saline lavage, intranasal or oral corticosteroids, and antibiotic therapy. If these or further medical therapies result in significant symptom improvement, patients may continue such treatments for the long-term to successfully manage their disease. However, many patients do not gain sufficient benefit from medical treatments and are offered the option of Endoscopic Sinus Surgery (ESS) \cite{Le2018}.

\subsection{Significance of CRS Surgical Outcome Prediction}

Analyses of the National Ambulatory Medical Care Survey (NAMCS) from the late 1980s and early 1990s reported rising office visits for sinusitis and identified sinusitis as one of the leading indications for antibiotic prescribing in U.S. outpatient practice \cite{Chow2012IDSA}. More recently, a national claims analysis showed that rhinosinusitis accounts for \emph{more outpatient antibiotic prescriptions than any other diagnosis}, underscoring its outsized contribution to community antibiotic use \cite{Smith2013JACI}. Chronic rhinosinusitis (CRS) is also a major economic burden: contemporary estimates place direct medical costs in the United States at roughly \$10–\$13\,billion annually, with an additional \(\sim\$20\,\)billion in indirect costs from productivity loss \cite{Rudmik2017,Smith2015Cost}. 

Given these statistics, it is paramount to achieve rapid and durable disease control in this chronic disease by quickly traversing the clinical management algorithm. In CRS, patient-reported quality of life (QoL) is a primary outcome and decision anchor, with instruments such as the SNOT‐22 recommended for tracking severity and response \cite{Hopkins2018CHROME,Orlandi2021ICAR,rudmik2015patient}. According to current estimates, about 300,000 ESS operations for CRS have been carried out annually in the U.S, with 30–40\% of patients showing unsatisfactory results in terms of achieving clinically significant SNOT-22 benefits \cite{Le2018,Soler2018Laryngoscope}. According to another recent study, 20-30\% of CRS patients who underwent endoscopic sinus surgery (ESS) reported that their post-treatment symptom reduction fell short of their expectations. Patient satisfaction with CRS intervention was strongly associated with the extent to which outcome expectations were met.
\cite{Mattos2019}. 

Most patients with medically refractory CRS do improve with surgery, but a substantial minority, on the order of one in three, derive limited benefit when SNOT-22 is the primary outcome of interest \cite{Le2018,Soler2018Laryngoscope}. This variability at the individual level motivates our study: if clinicians and patients could access \emph{pre-operative}, patient-specific estimates of surgical benefit, they could better weigh procedural and anesthesia risks against likely QoL gains, engage in more informed shared decision-making, and set realistic expectations \cite{Loftus2020JAMASurg}. Such individualized risk-benefit estimation is a central promise of personalized medicine and a key motivation for the rapid growth of structured data collection in healthcare over the past decade. Leveraging current AI/ML methods, we therefore investigate whether models trained on routinely available clinical data can estimate surgical outcomes to support counseling for elective ESS in CRS.

\section{Related \& Prior Work}

\subsection{Clinical guidance for CRS decision-making}
Contemporary care pathways for chronic rhinosinusitis (CRS) emphasize accurate phenotyping, optimization of medical therapy, and patient-centered shared decision-making about elective endoscopic sinus surgery (ESS) when symptoms remain refractory \cite{Orlandi2021ICAR,AAOHNSF2025Shin}. Patient-reported outcomes (PROs)--in particular the SNOT-22--are recommended to quantify baseline burden and to track response to therapy \cite{Hopkins2018CHROME,Orlandi2021ICAR}. Several longitudinal studies have demonstrated efficacy of ESS in this situation, and identified risk factors for treatment success/failure. But, the medical and sociodemographic variables are complex and interactive, and to date, no risk calculator exists in practice. Meta-analyses consistently show large average post-ESS improvements in SNOT-22, yet with wide interpatient variability \cite{Le2018,Soler2018Laryngoscope}. Current guidelines provide indication frameworks, but they stop short of offering individualized, pre-operative predictions of benefit using routinely available clinical variables \cite{AAOHNSF2025Shin}. This creates a practical gap for clinicians and patients seeking personalized risk–benefit estimates to guide shared decision-making.

\subsection{Predictive models in rhinology and sinus surgery}
Predictive modeling in rhinology remains comparatively sparse and heterogeneous. Prior efforts include regression-based and machine-learning (ML) approaches that associate baseline symptom burden and selected clinical variables with post-operative SNOT-22 trajectories or satisfaction; for example, Kang \emph{et~al.} used SNOT-22 to model outcomes after septoplasty \cite{Kang2020}, while large meta-analyses quantified average gains but were not designed as individualized risk calculators \cite{Le2018,Soler2018Laryngoscope,Mattos2019}. Much of the existing literature centers on population-level effect sizes, disease control categories, or biomarker/imaging correlates rather than point-of-care prediction from routine pre-operative clinical records \cite{Orlandi2021ICAR}. Consequently, there is no widely adopted, validated pre-operative tool that estimates a CRS patient’s probability of achieving a clinically meaningful SNOT-22 improvement after ESS using standard clinical data elements. A recent single-center study \cite{raghavan2025predicting} trained machine-learning models to predict attainment of SNOT-22 MCID after primary ESS using 59 preoperative predictors collected via an institutional survey platform, and reported strong discriminative performance for a stacking-style Ensemble model (AUC up to 0.89; accuracy 87.8\%) on an 80:20 split of 242 patients (test $n=48$). 
While promising, the study’s clinical generalizability remains uncertain because the cohort is drawn from a single academic sinus center and evaluation relies on a relatively small held-out test set. 
In addition, postoperative SNOT-22 outcomes were permitted as early as $\geq$2 months (when stability evidence is weaker than at $\geq$6 months), and the TabNet component required assigning specific numeric values to represent missing data, both of which may introduce additional variance and warrant careful sensitivity analysis. 
Overall, more rigorous validation (e.g., multicenter external validation and stronger deployment-oriented evaluation) is still needed before such models can be relied upon for point-of-care surgical counseling.

\subsection{Machine learning for surgical decision support}
Across surgery more broadly, interpretable ML is increasingly explored to augment case selection, perioperative risk stratification, and shared decision-making \cite{Loftus2020JAMASurg,Suh2022,Buch2018}. Also transparency in the process from data gathering, processing, assessment, experiments, reporting etc. is now expected, using TRIPOD+AI guidelines as we do here \cite{cohen2024tripod+} or similar alternative (CHAI \cite{CHAI_RAIG_2025}). A paper in this arena should follow at least one of these and mention which one. Best practices emphasize calibration, transparency of feature influence, and evaluation of net clinical utility versus treat-all or treat-none heuristics \cite{Loftus2020JAMASurg}. However, translation to CRS has lagged: most published clinical ML algorithms either rely on imaging-heavy pipelines or small, single-center cohorts, and they frequently omit core translational checks such as probability calibration and decision-curve (net-benefit) analysis—steps that are essential for clinical deployment \cite{Kang2020,Loftus2020JAMASurg,Collins2015TRIPOD,Vickers2006DCA,Steyerberg2019Book,Nagendran2020BMJ}.

\subsection{Gap addressed in this work}
In sum, while guidelines endorse PRO-driven management and meta-analyses demonstrate average post-ESS improvements, there remains a clear gap: a systematic ML application to \emph{routinely accessible} CRS clinical data that (i) frames the task as a pre-operative recommendation problem anchored to the SNOT-22 minimal clinically important difference, (ii) compares multiple algorithms and ensembles, (iii) prioritizes interpretability and visualization, and (iv) quantifies potential clinical utility compared to experts. Our study addresses these needs by building and evaluating such models on a multicenter CRS surgery cohort, and we report our approach and evaluation in a systematic fashion according to TRIPOD+AI recommendations.

\section{Dataset}
For predicting surgical outcomes (i.e., the potential effectiveness of ESS in CRS patients considering surgery), training the model is a major initial task.  Currently, the Sino-Nasal Outcome Test-22 (SNOT-22) is the most widely accepted sinus QOL tool for analyzing CRS patient-reported outcomes. It is an extensively validated set of 22 standard questions for understanding CRS symptom severity \cite{Le2018,Soler2018Laryngoscope,rudmik2015patient,Mattos2019}. Here, the patient rates each question on a Likert-type scale in the range of 0 to 5, resulting in a total score range between 0 to 110. Higher scores depict the worst QoL \cite{Soler2018Laryngoscope}. 

 To understand outcomes from medical and surgical intervention in CRS, an NIH-funded multicenter CRS outcomes study was designed and undertaken (clinicaltrials.gov NCT01332136). This study recorded detailed demographic and clinical metadata in a longitudinal fashion, along with patient-reported outcomes, including (1) a sinus-specific evaluation (SNOT-22), and (2) a general health measure (SF6d-derived health utility value, HUV) \cite{Brazier2002}. %and (3) olfactory outcomes. 
 A standardized collection of $\sim$30 clinical and demographic common data elements available to practicing physicians was performed throughout this study and used as the set of predictor variables.

We began with a development cohort of 791 patients (50 attributes; typical missingness for some clinical fields, referred to as the 2R01 dataset for the rest of the paper) and a second, independent cohort of 355 patients from a related multicenter CRS outcomes study with comparable clinician-recorded metadata (henceforth, referred to as the 3R01 dataset). In this paper, we restrict analysis to patients who actually underwent sinus surgery, because our objective is to build a model that flags cases that may be recommended to avoid surgery (i.e., would not achieve sufficient postoperative benefit). Operationally, this means we considered only the “surgery” cases, excluded patients managed medically (no surgery). So, we concatenated the two cohorts, and retained only those with sinus surgery as the treatment. After these filters, the combined dataset comprised 524 surgical patients. For this multicenter, observational CRS surgical outcomes
study with six different tertiary rhinology practice sites, institutional review board approval
was gained for each site, and de-identified data were shared for this investigation (COMIRB \#19-2085).

  The primary goal in this analysis is to be able to predict whether a surgery would result in the desired postoperative outcome (yes/no). To obtain that goal, we need to predict the change in total SNOT-22 from baseline to six months post-ESS, the point at which the treatment outcome is statistically considered stable and durable \cite{Kang2020}. The minimal clinically important difference (MCID) is a metric used to estimate the minimal amount of change in the SNOT-22 measurement that is noticeable clinically and is used to describe the clinical importance that results from a CRS intervention.  If the SNOT-22 score improvement is greater than or equal to an established MCID (8.9 points), it indicates that the intervention (ESS) resulted in a noticeable gain in sinus-related QoL \cite{Kang2020}. If the SNOT-22 score change was less than 8.9 points, then surgery did not result in noticeable benefit and perhaps continued medical therapy could have been continued unless there were other reasons to pursue surgical intervention. This categorization is used as a major binary output classification for the model training in the current paper.

\section{Methodology}

\subsection{Supervised learning paradigm}

Machine learning (ML) systems are commonly organized into three broad paradigms: \emph{supervised} learning, \emph{unsupervised} learning, and \emph{reinforcement} learning \cite{Bishop2006PRML,Hastie2009ESL,Goodfellow2016DL}. In supervised learning—the focus of this study—a model is learned from labeled examples \((x_i, y_i)\) to approximate an unknown mapping \(f:\mathcal{X}\!\to\!\mathcal{Y}\), such that
\[
y \approx f(x),
\]
where \(x\) denotes an input feature vector (e.g., demographics, comorbidities, baseline SNOT-22 items) and \(y\) is the associated target label (here, a binary indicator of surgical recommendation). The objective is not merely to fit the training data but to \emph{generalize} to unseen cases drawn from the same data-generating process \cite{Hastie2009ESL,Bishop2006PRML}.

Model development typically involves (i) splitting data into training/test sets or using cross-validation for hyperparameter tuning, (ii) fitting candidate algorithms, and (iii) assessing performance on the held-out test data with appropriate metrics \cite{Hastie2009ESL}. For binary clinical prediction tasks, accuracy alone can be misleading, particularly under class imbalance; complementary metrics such as precision, recall, F1-score, area under the ROC curve (AUROC), and confusion matrix are recommended \cite{Powers2011Eval,Steyerberg2019Book}. %Clinical translation further benefits from decision-curve (net-benefit) analysis to quantify potential utility across threshold preferences \cite{Vickers2006DCA}.

In this work, we framed CRS surgical selection as a supervised, pre-operative classification problem and compared multiple algorithms including regularized logistic regression, SVMs, random forests, gradient-boosted trees, Na{\"i}ve Bayes, a compact three-layer DNN, and a majority-vote ensemble, using stratified validation and a held-out test set. We report discrimination (e.g., AUROC), and error profiles (e.g., class-0 recall and macro-F1), consistent with best practices for clinical prediction modeling \cite{Steyerberg2019Book,Collins2015TRIPOD,Loftus2020JAMASurg}. Regarding patient and public involvement, neither was involved in the study design or any feedback collection.

\begin{figure}[t]
\centering
\includegraphics[width=\columnwidth]{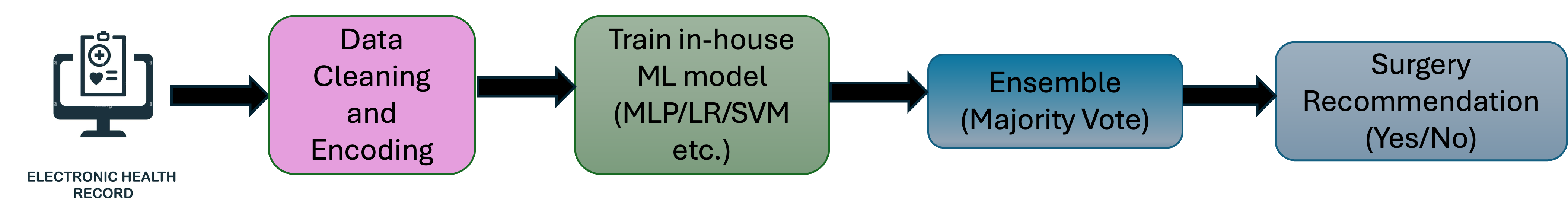}
\caption{Schematic pipeline of the proposed decision-support approach. Structured fields from the electronic health record (EHR) undergo data cleaning and encoding, after which multiple in-house classifiers (e.g., MLP, logistic regression, SVM) are trained. Model outputs are combined via a majority-vote ensemble to generate a binary surgery recommendation (Yes/No).}
\label{fig2}
\end{figure}

\subsection{Machine Learning Approaches Utilized}
\label{sec:methods}

We implement the binary classification models using \texttt{scikit-learn}~\cite{scikit-learn} and \texttt{XGBoost}~\cite{chen2016xgboost} libraries. 
In Stage~1, we train and benchmark individual classifiers (logistic regression, Na\"{i}ve Bayes, support vector machine, random forest, multi-layer perceptron, and XGBoost). 
In Stage~2, we form an ensemble by aggregating model outputs to improve predictive performance. 
Unless otherwise noted, hyperparameters were tuned via cross-validation, and performance was assessed on a held-out test set. The whole pipeline is schematically depicted in Fig.~\ref{fig2}. 

\subsubsection{Logistic Regression}
\label{subsec:logreg}
Logistic regression models the conditional probability of the positive class via the sigmoid of an affine function~\cite{bishop2006pattern}:
\begin{equation}
p(y=1 \mid \mathbf{x}) = \sigma(z), \qquad z = \mathbf{w}^{\top}\mathbf{x} + b, \qquad 
\sigma(z) = \frac{1}{1+\exp(-z)}.
\label{eq:logistic}
\end{equation}
Parameters $(\mathbf{w},b)$ are estimated by minimizing the regularized negative log-likelihood:
\begin{equation}
\min_{\mathbf{w},b} \; 
-\sum_{i=1}^{n} \big[ y_i \log \sigma(z_i) + (1-y_i)\log(1-\sigma(z_i)) \big]
\;+\; \lambda \, \mathcal{R}(\mathbf{w}),
\label{eq:logreg_obj}
\end{equation}
where $\mathcal{R}(\mathbf{w})$ is typically $\ell_2$ (ridge) or $\ell_1$ (lasso) regularization.

\subsubsection{Na\"{i}ve Bayes}
\label{subsec:nb}
Na\"{i}ve Bayes applies Bayes’ rule with a conditional independence assumption over features given the class~\cite{murphy2012ml}:
\begin{equation}
p(y \mid \mathbf{x}) \propto p(y) \prod_{j=1}^{d} p(x_j \mid y),
\label{eq:naivebayes}
\end{equation}
with the decision rule 
$\hat{y}=\arg\max_{y \in \{0,1\}} \left\{ \log p(y) + \sum_{j=1}^d \log p(x_j \mid y) \right\}$.
%We use Gaussian or multinomial likelihoods depending on feature type.

\subsection{Random Forest}
\label{subsec:rf}
Random forests are ensembles of decision trees trained on bootstrap samples with feature subsampling at each split~\cite{breiman2001random}. It creates decision trees from several observations. For classification, the forest prediction is the majority vote (or averaged class posterior) over $T$ trees $\{h_t\}_{t=1}^{T}$:
\begin{equation}
\hat{y} = \arg\max_{k \in \{0,1\}} \frac{1}{T} \sum_{t=1}^{T} \mathbb{I}\{h_t(\mathbf{x})=k\},
\label{eq:rf}
\end{equation}
and uses the average for regression problems.  In decision trees, each step involves a stingy selection of the optimal split point from the training dataset. By building several trees using various sampled datasets from the training datasets can decrease the high variance. This process is known as bootstrap aggregation, or bagging. 
Bagging reduces variance, and random feature selection decorrelates trees, improving generalization.

\subsection{Support Vector Machine (SVM)}
\label{subsec:svm}
This method is frequently applied for classification problems. SVM is effective in high-dimensional spaces. This algorithm projects each data sample as a point in an n-dimensional space, where n is the number of features. Next, classification is carried out by identifying the hyper-plane that effectively distinguishes the two classes. The selection of hyperplane is based on maximizing the distance between the closest data points of each class. SVM also uses different kernels, which can convert the low dimensional space into high dimensional spaces when the problem is non-linear.
For linearly separable data, the hard-margin SVM maximizes the geometric margin~\cite{cortes1995support}. In the soft-margin setting:
\begin{align}
\min_{\mathbf{w},b,\boldsymbol{\xi}} \quad & \frac{1}{2}\lVert \mathbf{w} \rVert_2^2 + C \sum_{i=1}^{n} \xi_i \\
\text{s.t.} \quad & y_i(\mathbf{w}^{\top}\phi(\mathbf{x}_i) + b) \ge 1 - \xi_i, \quad \xi_i \ge 0, \; i=1,\dots,n,
\label{eq:svm_primal}
\end{align}
where $\phi(\cdot)$ is an implicit feature map induced by kernel $K(\mathbf{x},\mathbf{x}')=\langle \phi(\mathbf{x}),\phi(\mathbf{x}') \rangle$. The decision function is 
\begin{equation}
f(\mathbf{x}) = \operatorname{sign}\!\left(\sum_{i=1}^{n} \alpha_i y_i K(\mathbf{x}_i,\mathbf{x}) + b\right),
\label{eq:svm_dual}
\end{equation}
with dual coefficients $\alpha_i$ found by quadratic programming.

\subsection{Multi-Layer Perceptron (MLP)}
\label{subsec:mlp}
Usually, Artificial Neural Networks are referred to as neural networks or Multi-Layer Perceptrons since it is the most widely used architecture in the field of neural networks. A perceptron incorporates a single neuron model, which is a node in larger neural networks. The layered structure of the framework of the network gives the power of prediction for neural networks. The layered architecture consists of the input layer, hidden layers, and an output layer. 
An $L$-layer MLP composes affine maps and nonlinearities~\cite{goodfellow2016deep}:
\begin{equation}
\begin{aligned}
\mathbf{h}^{(0)}&=\mathbf{x},\quad 
\mathbf{h}^{(\ell)}=\sigma^{(\ell)}\!\left(\mathbf{W}^{(\ell)}\mathbf{h}^{(\ell-1)}+\mathbf{b}^{(\ell)}\right),\;
\ell=1,\dots,L-1,\\
\hat{p}&=\mathrm{sigmoid}\!\left(\mathbf{w}^{\top}\mathbf{h}^{(L-1)}+b\right).
\end{aligned}
\label{eq:mlp_forward}
\end{equation}
trained by minimizing binary cross-entropy (optionally with weight decay) via stochastic gradient descent, with the loss function being:
\begin{equation}
\mathcal{L} = -\frac{1}{n}\sum_{i=1}^{n} \left[ y_i \log \hat{p}_i + (1-y_i)\log(1-\hat{p}_i) \right] + \lambda \sum_{\ell}\lVert \mathbf{W}^{(\ell)} \rVert_2^2.
\label{eq:mlp_loss}
\end{equation}

\subsection{Extreme Gradient Boosting (XGBoost)}
\label{subsec:xgb}
XGBoost builds an additive ensemble of regression trees to minimize a regularized objective using second-order (Newton) boosting~\cite{chen2016xgboost,friedman2001greedy}. 
Let $F_M(\mathbf{x})=\sum_{m=1}^{M} f_m(\mathbf{x})$ be the model, where each $f_m$ is a tree. The objective is
\begin{equation}
\mathcal{L}^{(M)} 
= \sum_{i=1}^{n} \ell\!\left(y_i, \hat{y}_i^{(M-1)} + f_M(\mathbf{x}_i)\right) 
\;+\; \Omega(f_M),
\qquad 
\Omega(f) = \gamma T + \frac{\lambda}{2}\lVert \mathbf{w} \rVert_2^2,
\label{eq:xgb_obj}
\end{equation}
where $T$ is the number of leaves, $\mathbf{w}$ are leaf scores, and $\ell$ is the logistic loss for binary classification. 
Using a second-order Taylor expansion around $\hat{y}_i^{(M-1)}$, the split gain for a candidate partition $(L,R)$ with gradients $g_i=\partial_{\hat{y}}\ell$ and Hessians $h_i=\partial^2_{\hat{y}}\ell$ is
\begin{equation}
\mathrm{Gain} 
= \frac{1}{2} \left( 
\frac{\left(\sum_{i \in L} g_i\right)^2}{\sum_{i \in L} h_i + \lambda} 
+ 
\frac{\left(\sum_{i \in R} g_i\right)^2}{\sum_{i \in R} h_i + \lambda} 
- 
\frac{\left(\sum_{i \in L \cup R} g_i\right)^2}{\sum_{i \in L \cup R} h_i + \lambda}
\right) - \gamma.
\label{eq:xgb_gain}
\end{equation}
Trees are added greedily to maximize \eqref{eq:xgb_gain} until convergence or early-stopping criteria are met.

\begin{figure}[t]
\centering
\includegraphics[width=\columnwidth]{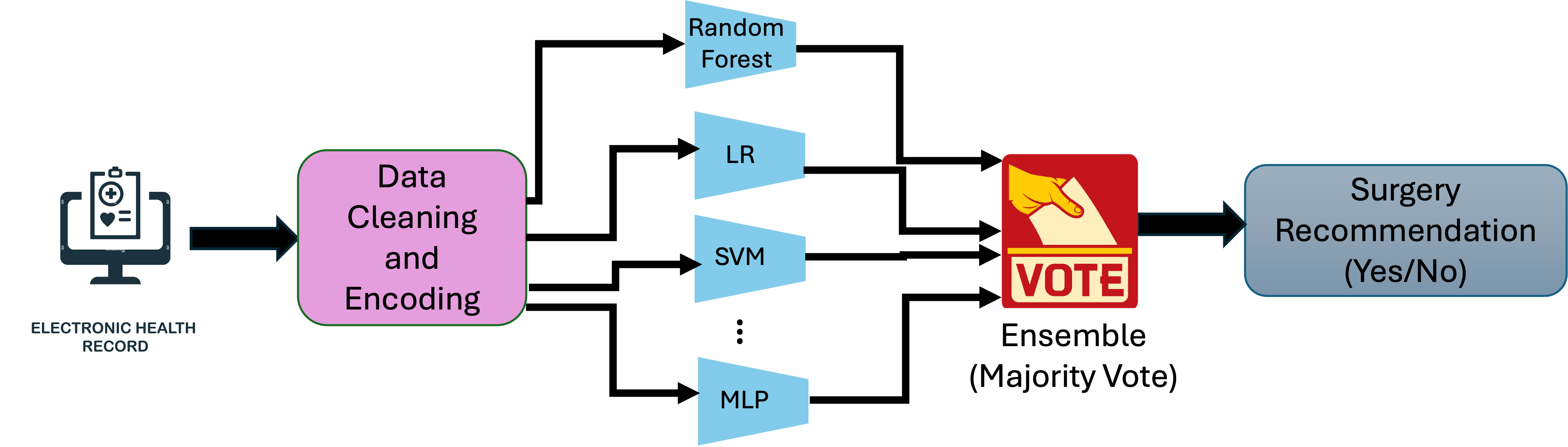}
\caption{Ensemble decision-support pipeline. Structured EHR data are cleaned and encoded, then passed in parallel to multiple base learners (\texttt{Random Forest}, \texttt{Logistic Regression}, \texttt{SVM}, \texttt{MLP} etc.). Their predictions are aggregated via a voting scheme to produce a binary surgery recommendation (Yes/No). Only preoperative variables are used for inference.}
\label{fig3}
\end{figure}

\paragraph{Ensemble aggregation.}
Given $M$ trained models producing class posteriors $p_m(y=k \mid \mathbf{x})$, we use soft voting with nonnegative weights $\{w_m\}_{m=1}^M$ (normalized to sum to 1):
\begin{equation}
\hat{y} = \arg\max_{k \in \{0,1\}} \sum_{m=1}^{M} w_m \, p_m(y=k \mid \mathbf{x}).
\label{eq:softvote}
\end{equation}
We also experimented with stacking (logistic meta-learner) using out-of-fold predictions as meta-features~\cite{wolpert1992stacked}. In particular, the majority voting ensemble process during inference is illustrated schematically in Fig.~\ref{fig3}. As shown in Fig.~\ref{fig3}, the pipeline ingests structured clinical variables, performs data cleaning and encoding, and feeds the resulting features in parallel to multiple base learners (\texttt{Random Forest}, \texttt{Logistic Regression}, \texttt{SVM}, \texttt{MLP} etc.). Their predictions are aggregated via a voting scheme to yield a binary surgery recommendation (Yes/No). Only preoperative variables are used as features for training to prevent information leakage; the ensemble reduces variance and mitigates model-specific biases, providing more stable recommendations than any single classifier.

\section{Dataset Preparation}
\subsection{Cleaning and Encoding}
Dataset cleaning is an important step for every machine learning model. This is of particular relevance for clinical data, as there are often challenges in the systematic measurement and recording of data that result from varied human factors. A messy dataset might negatively impact the prediction model, so it is necessary to examine and understand any flaws and address them wherever possible before training a model alongside reporting this process transparently. Here, this was accomplished in two parts. First, the data were cleaned and then preprocessed (encoding, then categorical to numerical conversion) to obtain a set of features suitable for ML training. Note, this process was repeated for both the 2R01 and 3R01 datasets.

For the 2R01 dataset, initially, there were a  total of 791 rows of patient data with 50 attributes as columns. First, we chose the subset where treatment was sinus surgery, reducing the dataset to 604 rows. There were missing data and unwanted attributes present which might not be useful for accurate surgery outcome prediction. So, in the cleaning process,  all the unwanted columns were removed from this dataset under a physician’s guidance. Then, all the blank spaces (i.e., the null values) in all the columns of the patient were removed. Then the total null value counts were found  as shown in Table \ref{tab:nulls_2r01_surgery}.

\begin{table}[t]
\centering
\caption{Feature-wise missing (null) values in the 2R01 cohort restricted to surgery-treated patients.}
\label{tab:nulls_2r01_surgery}
\begin{tabular}{l r}
\toprule
\textbf{Feature} & \textbf{\# Nulls} \\
\midrule
Age                      & 1 \\
Race                     & 1 \\
Education                & 10 \\
Household Income        & 7 \\
Smoker                  & 4 \\
Alcohol                  & 5 \\
Diabetes                 & 1 \\
Baseline CT Score           & 2 \\
Baseline Endoscopy Score    & 3 \\
SNOT-22 Baseline Score       & 1 \\
%SNOT-22 Score After 6 Months & 217 \\
\bottomrule
\end{tabular}
\end{table}

To establish the target outcome, the difference between Baseline SNOT-22 and 6-month postoperative SNOT-22 scores was calculated for each individual. After removing the 217 patient samples where patients did not come back for followup after 6 months (as such, we do not have a label for them), baseline Health Utility score columns were also removed since the study concentrated on the SNOT-22 score. Now to find the difference between the baseline SNOT-22 score and the SNOT-22 score after 6 months of operation, we changed their datatype into integers since the values in those columns were object types. Then, the differences between those two columns were calculated and assigned as the target column. Additionally, the 6 month SNOT-22 score column was removed as it is a post-operative  attribute. Also, the rest of the null value rows (Table 1) were removed as they were very few in number. Then the dimensions of the dataset became (371,31). Overall, the missingness of the fields in the dataset was 4.7\%, which is quite low. We also experimented with data imputation for the missing values, however, as these cases were very few (3\% only), it did not affect the results compared to excluding the subject (sample) altogether. Similar data cleaning was performed on the 3R01 dataset, which initially had 354 rows of patient data with 39 attributes as columns. Zooming in on just the surgery treatment cases resulted in 266 rows. The null values in specific columns are shown in Table \ref{tab:nulls_3r01_surgery}. Similar operation was carried out to create the targets, which resulted in a cleaned dataset with dimensions (153,31). The attributes for this 3R01 dataset were organized to overlap with those in the 2R01 dataset. As in the 2R01 dataset case, the target column for the 3R01 dataset was also changed from continuous values to binary outcomes based on the established SNOT-22 8.9 point MCID. %As mentioned earlier, due to the relatively small dataset size, we concatenated the cleaned 2R01 and 3R01 datasets to obtain a combined dataset, the following discussions throughout the rest of the paper will be based on this dataset.

\begin{table}[t]
\centering
\caption{Feature-wise missing (null) values in the 3R01 cohort restricted to surgery-treated patients.}
\label{tab:nulls_3r01_surgery}
\begin{tabular}{l r}
\toprule
\textbf{Feature} & \textbf{\# Nulls} \\
\midrule
Race                       & 2 \\
Education                  & 3 \\
Household Income          & 1 \\
Obstructive Sleep Apnea (OSA)  History               & 3 \\
Alcohol                   & 3 \\
Baseline Endoscopy Score      & 3 \\
%BSIT\_DIAGNOSIS\_string    & 266 \\
%BLN\_FULL\_SIT40\_SCORE    & 266 \\
%BLN\_BSIT\_SCORE           & 266 \\
%BSIT\_6MONTH               & 266 \\
SNOT-22 Baseline Score         & 3 \\
%SNOT-22 Score After 6 Months & 105 \\
%SF6D\_HUV\_BLN             & 3 \\
%SF6D\_SCORE\_6MONTH        & 104 \\
\bottomrule
\end{tabular}
\end{table}

\begin{figure}[t]
\centering
\includegraphics[width=\columnwidth]{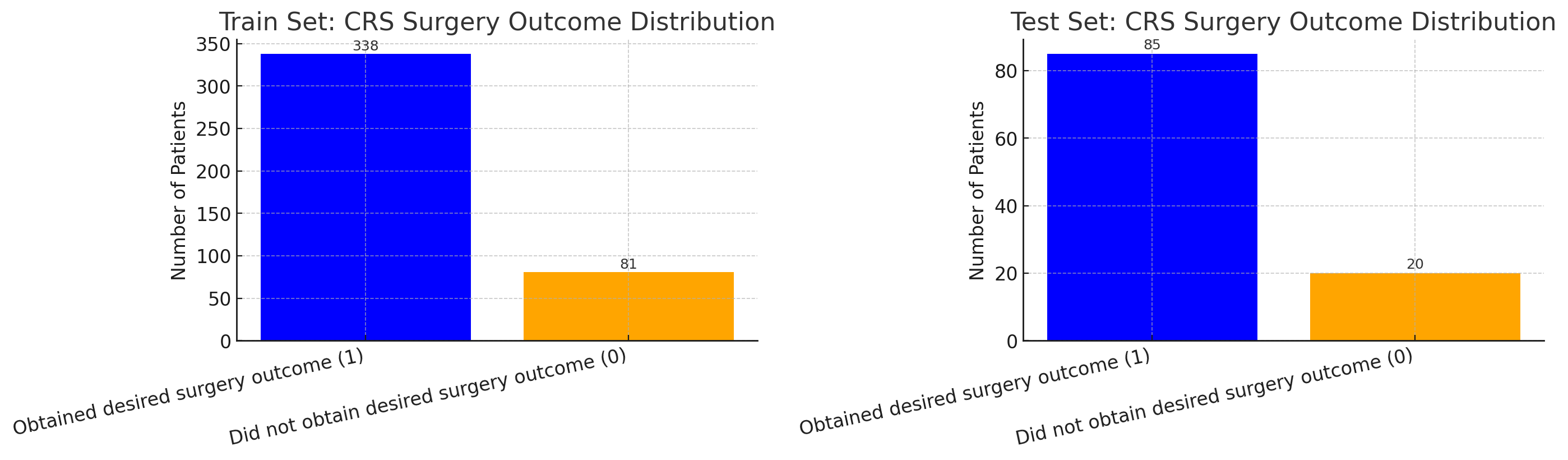}
\caption{Class distribution in the train (left) and test (right) sets. Bars show the number of CRS patients who obtained the desired surgery outcome (1) versus those who did not obtain the desired outcome (0). The splits retain the overall class imbalance (train: 338 vs. 81; test: 85 vs. 20), which is considered in model training and evaluation.}
\label{fig:Class Distribution Bar Chart}
\end{figure}

Next, the categorical variables were converted to numeric codes using deterministic, clinically sensible mappings. This encoding is necessary because most machine-learning algorithms operate on numerical tensors and require consistent, reproducible representations of categorical information to learn stable decision boundaries. Specifically, demographic/socioeconomic fields (e.g., \texttt{Sex}, \texttt{Race}, \texttt{Ethnicity}, \texttt{Education}, \texttt{Household Income}, \texttt{Insurance}) and clinical history variables (e.g., \texttt{Previous Surgery}, \texttt{Positive allergy testing }, \texttt{Smoker}, \texttt{Alcohol}, \texttt{Chronic Obstructive Pulmonary Disease \newline(COPD)}, \texttt{Aspirin Intolerance}, \texttt{Obstructive Sleep Apnea (OSA) History}, \texttt{Gastroesophageal Reflux Disease (GERD)}, \texttt{Allergic Fungal Sinusitis (AFS)}, \texttt{Asthma}, \texttt{Diabetes}, \texttt{Fibromyalgia}) were recoded from text labels into integer categories via explicit dictionaries (for example, \emph{Female}$\rightarrow$0, \emph{Male}$\rightarrow$1 for \texttt{Sex}; \emph{Employer provided}$\rightarrow$0, \emph{Medicare}$\rightarrow$1, \emph{Private}$\rightarrow$2, \emph{Canadian Medicare}$\rightarrow$3, \emph{Medicaid}$\rightarrow$4 for \texttt{Insurance}). Continuous baseline measures, such as the \texttt{SNOT-22 Baseline Score}, \texttt{Baseline CT Score}, and \texttt{Baseline Endoscopy Score}, were retained as numeric features. Where applicable, missing values were handled using simple domain-consistent rules (e.g., mapping textual ``None'' to a sentinel category for certain comorbidities); otherwise they were left for model-specific handling. After the encoding, we verified type consistency, and exported the encoded dataset for downstream train/test splitting and model fitting. This encoding scheme enables reproducible training across classifiers and avoids leakage from post-treatment or follow-up fields.

Next, we partitioned the dataset into an \textbf{80:20 train--test split} using stratified sampling on the binary outcome to preserve class prevalence. \textbf{Stratified sampling} here means that we performed the split \emph{within each outcome class} (class~1 and class~0) and then combined the selected patients, yielding similar class proportions in both subsets. The training set comprised 338 patients with the desired postoperative outcome (class~1) and 81 without (class~0); the test set comprised 85 class~1 and 20 class~0 patients. This yields closely matched prevalence across splits (train: 80.6\% vs.\ 19.3\%; test: 81.0\% vs.\ 19.0\%). As shown in Fig.~\ref{fig:Class Distribution Bar Chart}, the side-by-side bar plots depict these counts, with blue indicating patients who obtained the desired outcome (1) and orange indicating those who did not (0). Stratification enables fair comparison between training and held-out performance.

\subsection{Feature Correlation Analysis}
\begin{figure}[t]
\centering
\includegraphics[width=\columnwidth]{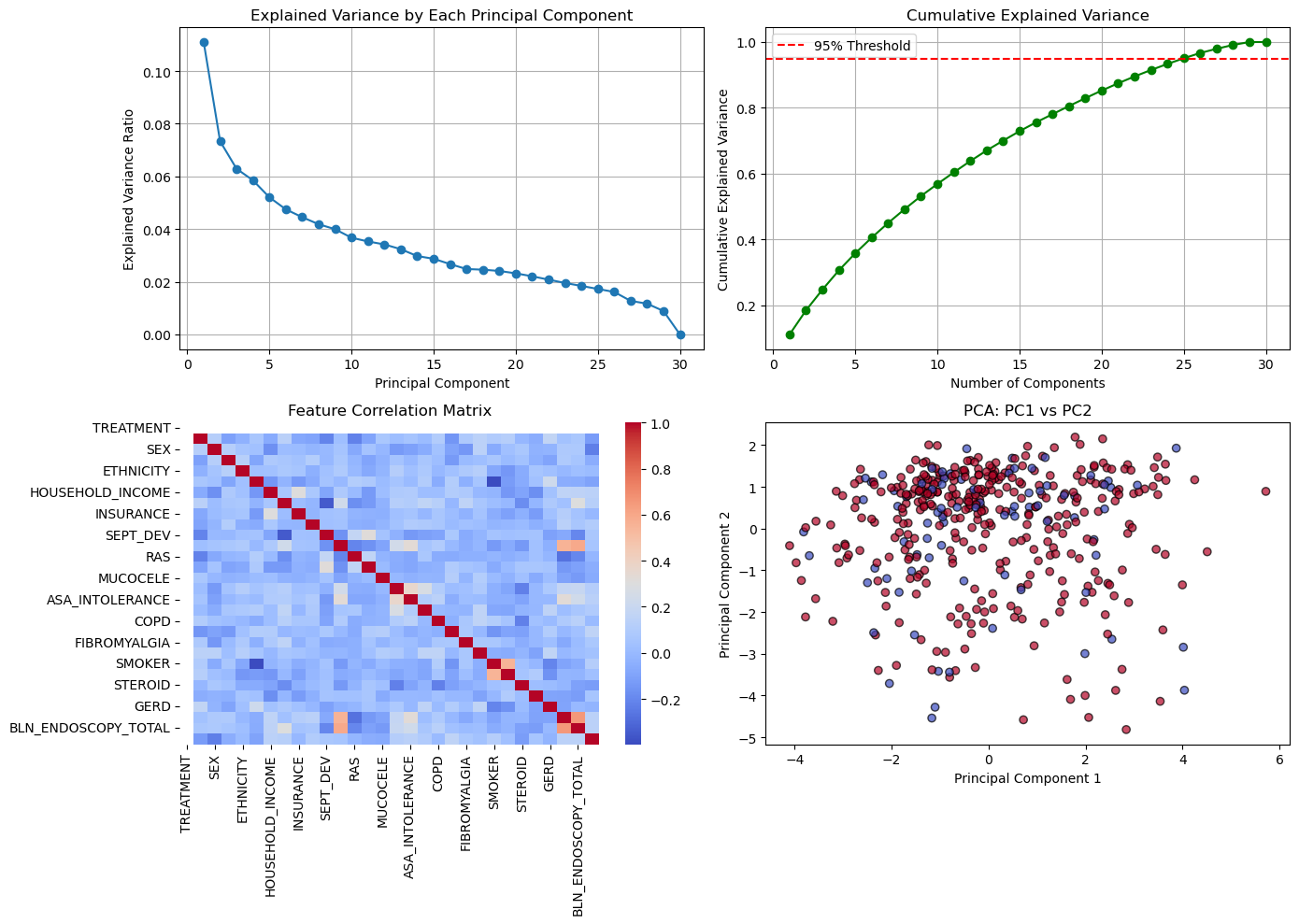}
\caption{Exploratory feature and PCA analysis.
(a) Explained variance by principal component. Each bar shows the proportion of total variance attributable to an individual component after standardization; early components carry the largest share, followed by a long tail.
(b) Cumulative explained variance. The running total of variance captured as components are added; the dashed line marks 95\% retained variance, indicating that a moderate subset of components suffices for compact representations.
(c) Feature correlation heatmap (Pearson’s correlation coefficient, $r$). Blue denotes negative and red positive associations among preprocessed clinical predictors; the matrix helps flag redundancy/collinearity before modeling.
(d) PCA score plot (PC1 vs. PC2). Patients are projected onto the first two principal components and colored by class label; the overlap across classes in this linear 2D view suggests that discrimination likely depends on higher-order components and/or non-linear structure.}
\label{fig:feature_correlation}
\end{figure}
To explore linear structure and potential dimensionality reduction, we applied Principal Component Analysis (PCA) to the standardized feature matrix (zero mean, unit variance; no outcome or post-treatment variables included). The explained-variance plot (top left in Fig.~\ref{fig:feature_correlation}) shows that the first few components account for a disproportionate share of variance, followed by a long tail of smaller contributions. The cumulative explained-variance curve (top right in Fig.~\ref{fig:feature_correlation}) indicates that retaining a moderate number of principal components captures the vast majority of variance (the 95\% reference line is shown), which is useful for building compact visualizations or low-dimensional baselines while avoiding information leakage. Finally, the PC1–PC2 scatter (colored by class) reveals substantial overlap between groups in the first two linear components (bottom right in Fig.~\ref{fig:feature_correlation}), suggesting that class separation, if present, likely resides in higher-order directions and/or is non-linear. This motivates the use of flexible learners (e.g., tree ensembles, kernels, or neural networks) and careful regularization rather than relying solely on linear boundaries in a low-dimensional subspace.
Note, PCA here is used for exploratory analysis and visualization; model training uses the original engineered/encoded features with appropriate preprocessing. Furthermore, we examined pairwise linear associations among all preprocessed clinical predictors using the Pearson correlation coefficient after encoding categorical variables and standardizing continuous features, as shown in bottom left plot of Fig.~\ref{fig:feature_correlation}. The heatmap (blue = negative, red = positive) visualizes the full correlation matrix, with the diagonal fixed at 1 and off-diagonal cells indicating the strength and direction of association between feature pairs. This screening step helps identify redundancies and potential collinearity prior to modeling and interpretation. While tree-based learners (e.g., Random Forest, XGBoost) are relatively robust to correlated inputs, we still monitor high absolute correlations to avoid unstable coefficients in linear models (e.g., logistic regression) and to interpret variable importance more cautiously. No outcome or post-treatment variables were included in the matrix to prevent leakage.

\section{Results and Analysis}
\subsection*{First Level Results}
In the first level, the training split of the combined dataset was used to individually train the following classifiers: Logistic Regression, Multi-Layer Perceptron (MLP), Random Forest (RF), Support Vector Machine (SVM,) and Naïve Bayes. Then, the testing split (20\% of data) was used for evaluating the trained model. Note, once the models were trained, they were not updated based on new data. The results are presented below: 

\begin{figure}
\centering
\includegraphics[width=.9\columnwidth, height=8in]{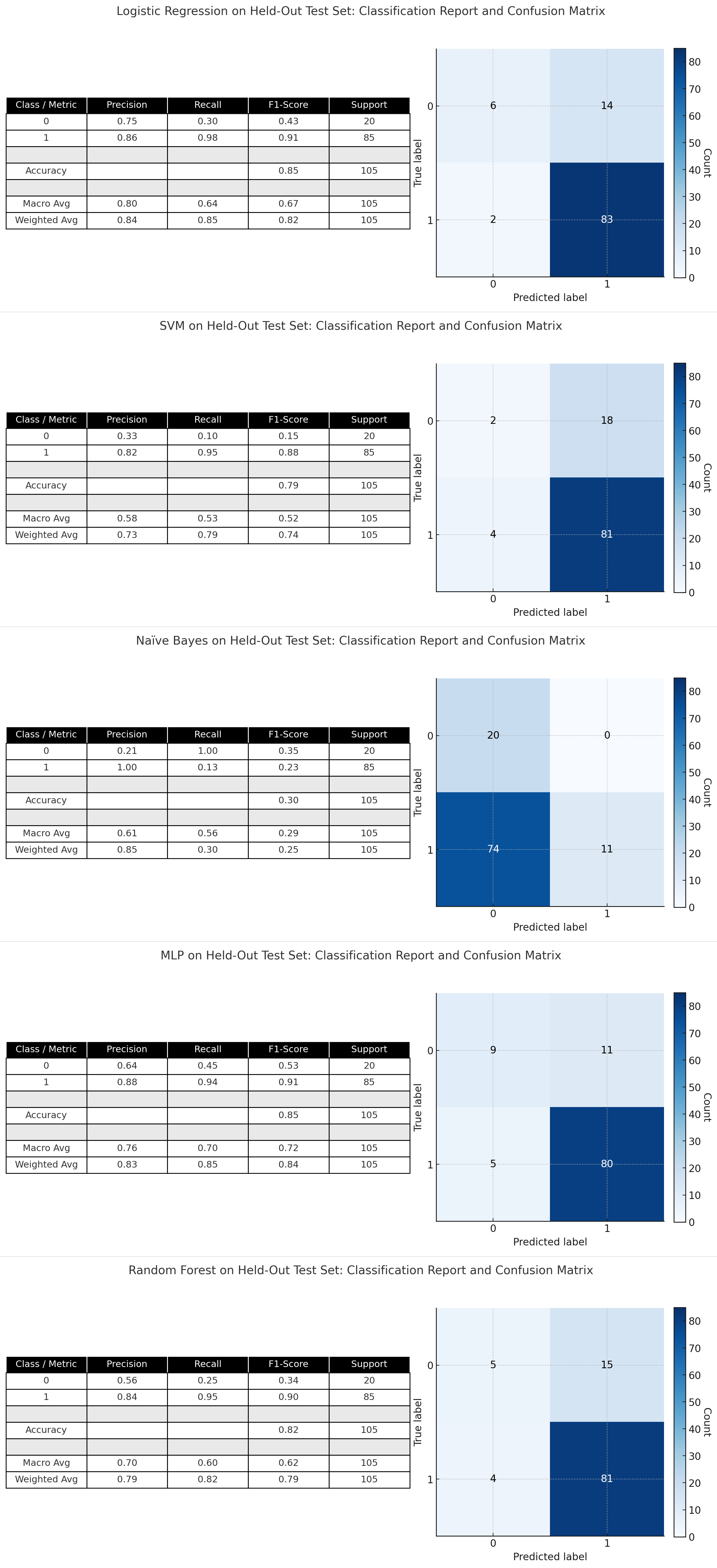}
\caption{Held-out test performance across five classifiers (Logistic Regression, SVM, Na\"{i}ve Bayes, MLP, Random Forest). Each panel shows the classification report (precision, recall, F1-score, support) alongside the confusion matrix (counts).}
\label{fig:model_results_bundle}
\end{figure}

\subsubsection{Logistic Regression}
As shown in the \textbf{top row} of Fig.~\ref{fig:model_results_bundle}, logistic regression attains overall accuracy \textbf{0.85}. Class~1 (desired outcome) performance is strong (precision \textbf{0.86}, recall \textbf{0.98}, F1 \textbf{0.91}), while class~0 remains challenging (precision 0.75, recall 0.30, F1 0.43). The confusion matrix \([6,14; 2,83]\) indicates a tendency to predict class~1, prioritizing sensitivity over specificity.

\subsubsection{Support Vector Machine (SVM)}
As shown in the \textbf{second row} of Fig.~\ref{fig:model_results_bundle}, SVM achieves accuracy \textbf{0.79}. It performs well on class~1 (precision \textbf{0.82}, recall \textbf{0.95}, F1 \textbf{0.88}) but exhibits low recall for class~0 (0.10; F1 0.15). The confusion matrix \([2,18; 4,81]\) reflects many class~0$\rightarrow$1 errors, consistent with the class imbalance.

\subsubsection{Na\"{i}ve Bayes}
As shown in the \textbf{third row} of Fig.~\ref{fig:model_results_bundle}, Na\"{i}ve Bayes underperforms with accuracy \textbf{0.30}. It perfectly recalls class~0 (precision 0.21, recall \textbf{1.00}, F1 0.35) but misses most positives (class~1 recall 0.13, F1 0.23). The confusion matrix \([20,0; 74,11]\) suggests the conditional independence assumptions are poorly matched to this dataset.

\subsubsection{Multi-Layer Perceptron (MLP)}
As shown in the \textbf{fourth row} of Fig.~\ref{fig:model_results_bundle}, the MLP reaches accuracy \textbf{0.85}. It balances precision and recall for class~1 (precision \textbf{0.88}, recall \textbf{0.94}, F1 \textbf{0.91}) and improves class~0 detection relative to linear/kernal baselines (recall 0.45, F1 0.53). The confusion matrix \([9,11; 5,80]\) indicates better control of false positives than SVM.

\subsubsection{Random Forest}
As shown in the \textbf{bottom row} of Fig.~\ref{fig:model_results_bundle}, Random Forest attains accuracy \textbf{0.82}. Positive-class performance is strong (precision \textbf{0.84}, recall \textbf{0.95}, F1 \textbf{0.90}), while class~0 recall remains modest (0.25; F1 0.34). The confusion matrix \([5,15; 4,81]\) highlights a minority-class miss pattern similar to other models, though tree ensembling yields robust overall discrimination.

\begin{figure}
\centering
\includegraphics[width=.9\columnwidth]{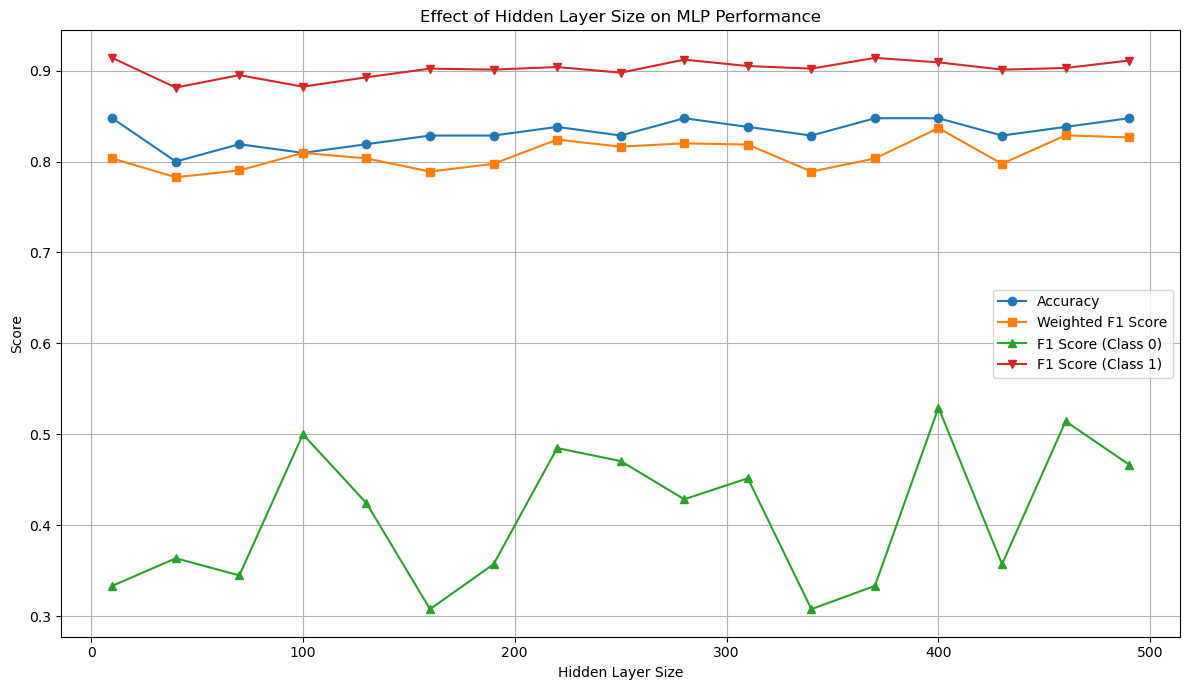}
\caption{MLP hyperparameter tuning (single hidden layer). Validation performance vs.\ hidden-layer width showing accuracy, weighted F1, and class-specific F1 scores. Overall accuracy and class~1 F1 remain relatively stable across widths, whereas the class~0 F1 (our priority) improves with larger layers and peaks for wide settings (\(\sim\)400--480 units). Results are from stratified cross-validation on the training split.}

\label{fig:mlp_tuning}
\end{figure}

\begin{figure}
\centering
\includegraphics[width=.9\columnwidth]{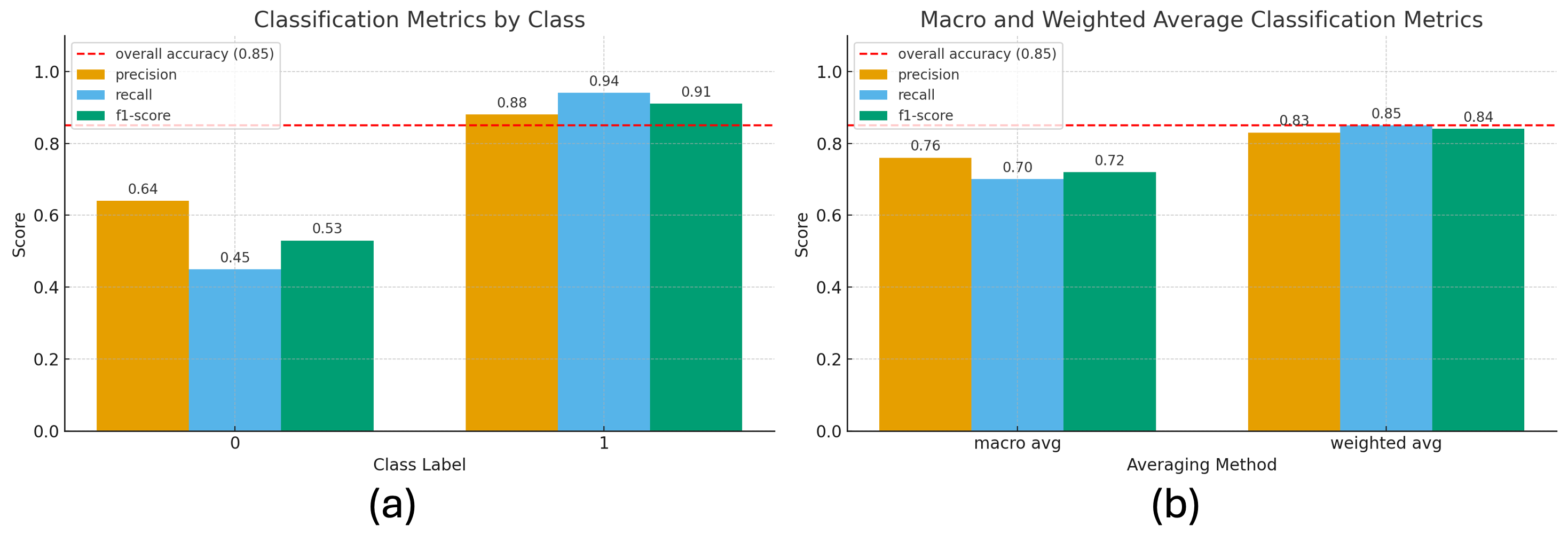}
\caption{Optimized MLP performance on the held-out test set. Left: class-wise precision, recall, and F1 for class~0 and class~1 (dashed line marks overall accuracy, 0.85). Right: macro- and weighted-average metrics, which are close to overall accuracy, indicating stable performance without overfitting to the majority class.}
\label{fig:mlp_classwise_results}
\end{figure}

\begin{figure}
\centering
\includegraphics[width=.9\columnwidth]{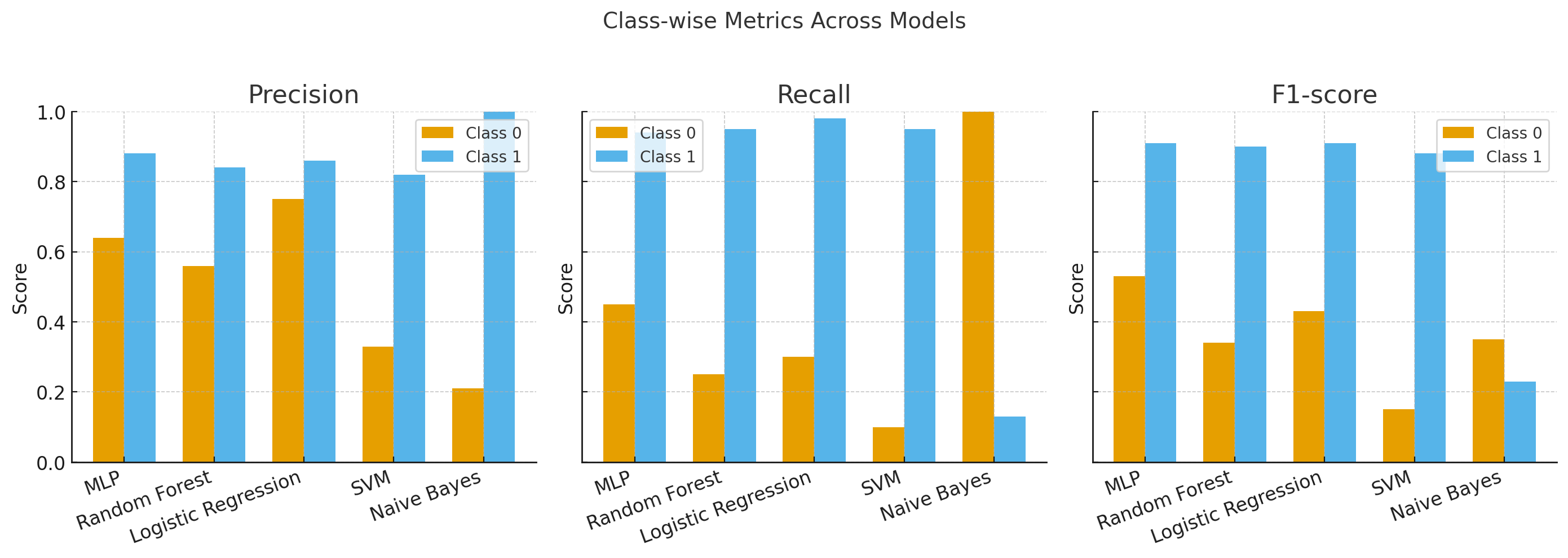}
\caption{Class-wise comparison of precision, recall, and F1-score across five models on the held-out test set (MLP, Random Forest, Logistic Regression, SVM, Na\"{i}ve Bayes). Each panel shows grouped bars for class~0 vs.\ class~1. Overall, class~1 metrics are higher for all models, while the MLP attains the strongest class~0 F1 among the candidates.}
\label{fig:classwise_metrics_across_models}
\end{figure}

\subsubsection{Hyperparameter tuning of the MLP}
All candidate models were hyperparameter optimized via stratified cross-validation on the training split. Because the MLP achieved the best trade-off for our primary goal (detecting class~0), we report its tuning here. We first varied network depth and observed that architectures with more than one hidden layer consistently underperformed a single-hidden-layer model, particularly on the class~0 F1. Given the limited dataset size, this behavior is consistent with overfitting from excess capacity. We therefore fixed a \emph{single hidden layer} and tuned its width. As shown in Fig.~\ref{fig:mlp_tuning}, overall accuracy and weighted F1 were relatively stable (\(\approx 0.82\text{--}0.85\)) across widths, whereas the \emph{class~0 F1} varied more markedly, improving with larger widths and peaking for wide layers (e.g., \(\sim\!400\text{--}480\) units). The class~1 F1 remained high (\(\approx 0.88\text{--}0.92\)) throughout. We selected a single-hidden-layer MLP with 400 neurons, combined with early stopping, class weighting, and \(L_2\) regularization, to maximize class~0 detection while maintaining strong overall performance. Figure~\ref{fig:mlp_classwise_results} summarizes the optimized MLP’s performance on the held-out test set. 
Class-wise metrics (left) highlight the asymmetry between classes - high precision/recall for class~1 
(\(F_{1}=0.91\)) and moderate but clinically useful performance for class~0 (\(F_{1}=0.53\)), while the 
macro/weighted averages (right) remain close to the overall accuracy (\(0.85\)), indicating stable 
behavior without excessive overfitting to the majority class.

\subsubsection{Comparison among the Models}
Because our clinical objective is to \emph{prospectively flag patients unlikely to achieve the desired postoperative outcome} (class~0), we prioritize models with strong class~0 detection rather than overall accuracy alone. As shown in Fig.~\ref{fig:classwise_metrics_across_models}, the MLP offers the best trade-off for this goal: it attains the highest class~0 F1-score (\textbf{0.53}) with improved recall (\textbf{0.45}) relative to logistic regression (F1 0.43, recall 0.30), SVM (F1 0.15, recall 0.10), and random forest (F1 0.34, recall 0.25), while maintaining strong performance on class~1 (F1 0.91) and high overall accuracy (0.85). Although Na\"{i}ve Bayes recalls all class~0 cases (recall 1.00), it severely underperforms on class~1 (recall 0.13) and overall accuracy (0.30), making it unsuitable for balanced clinical use. Accordingly, the MLP is the preferred model for downstream analyses and prospective decision support.

\subsubsection{XGBoost}
\begin{figure}[t]
  \centering
  \includegraphics[width=0.95\linewidth]{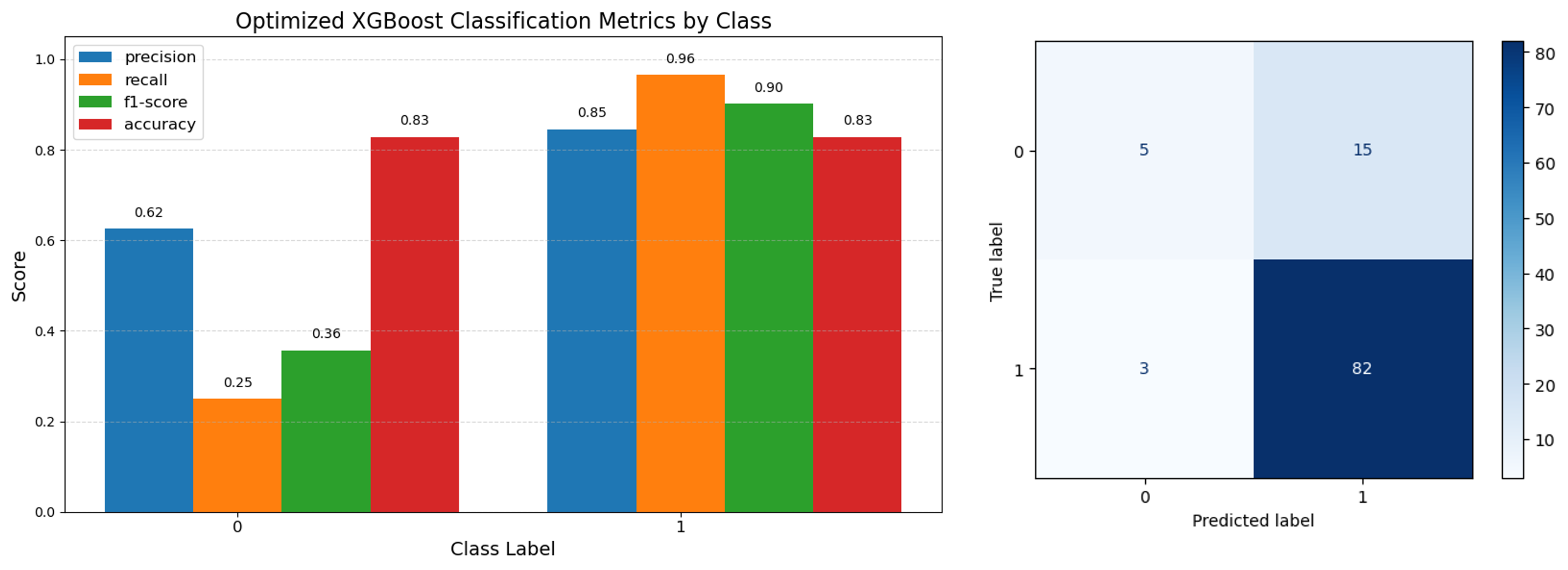}
  \caption{Optimized XGBoost results on the held-out test set. Left: class-wise precision, recall, \(F_1\), and overall accuracy bars (class~0 vs.\ class~1). Right: confusion matrix (counts), yielding accuracy \(\approx 0.83\) with strong class~1 recall (\(0.96\)) but modest class~0 recall (\(0.25\)).}
  \label{fig:xgb_results}
\end{figure}

We trained gradient-boosted decision trees (XGBoost) and tuned hyperparameters via a stratified grid search on the training split. The optimal configuration was:
\(\{\texttt{colsample\_bytree}=1.0,\ \texttt{learning\_rate}=0.05,\ \texttt{max\_depth}=3,\ \texttt{n\_estimators}=200,\ \texttt{subsample}=0.8\}\).
As summarized in Fig.~\ref{fig:xgb_results}, the tuned model achieved overall accuracy \(\approx 0.83\) on the held-out test set with confusion matrix
\(\begin{bmatrix}5 & 15\\ 3 & 82\end{bmatrix}\).
Class-wise metrics show strong performance for class~1 (precision \(0.85\), recall \(0.96\), \(F_1=0.90\)), but more limited detection of class~0 (precision \(0.62\), recall \(0.25\), \(F_1=0.36\)).
Because our clinical objective prioritizes identifying class~0 patients, the XGBoost model underperforms the MLP, which attains a higher class~0 \(F_1\) (\(0.53\)) while maintaining comparable overall accuracy (\(0.85\)). We therefore retain the MLP as the preferred model for downstream analysis and decision support.

\subsection*{Second Level Results: The Ensemble Approach}
Combining complementary classifiers in an ensemble can improve generalization by averaging out model–specific errors and leveraging diversity among base learners, often outperforming any single model. Motivated by this, we implemented and evaluated \emph{three} ensemble strategies, described in the next subsections, to assess whether aggregating heterogeneous learners yields more reliable predictions for our task.

\subsubsection{Majority-voting ensemble}
To assess whether combining complementary inductive biases could further stabilize predictions, we implemented a simple \emph{hard-voting} ensemble using six independently tuned base learners: Logistic Regression, SVM, Random Forest, Na\"{i}ve Bayes, MLP, and XGBoost (see individual results above). For each test instance \(x_i\), the ensemble aggregates the class labels \(\hat{y}^{(m)}_i \in \{0,1\}\) from model \(m=1,\dots,6\) via majority vote
\[
\hat{y}^{(\mathrm{ens})}_i \;=\;
\arg\max_{c\in\{0,1\}}
\sum_{m=1}^{6}\mathbf{1}\!\left(\hat{y}^{(m)}_i = c\right),
\]
with a deterministic tie–break in favor of the \textbf{MLP} prediction (consistent with our objective of maximizing class~0 detection and with its strong standalone performance).

\begin{figure}[t]
  \centering
  \includegraphics[width=0.95\linewidth]{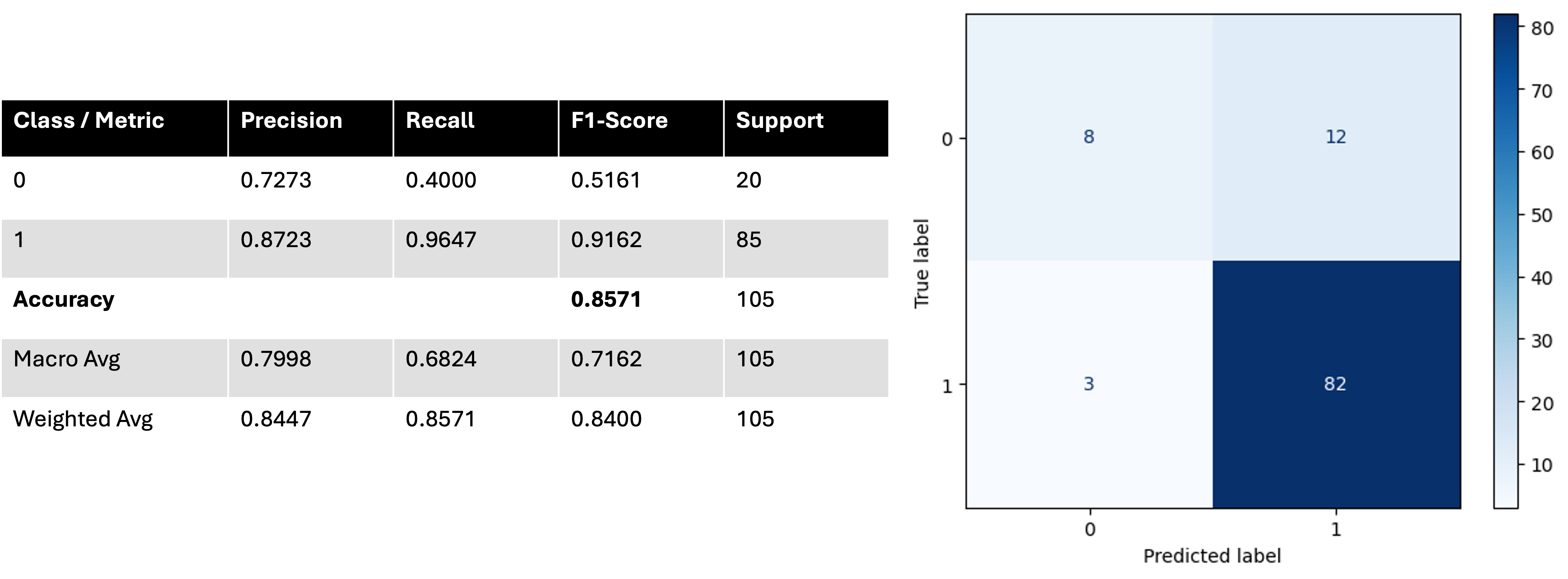}
  \caption{Majority-voting ensemble on the held-out test set. Left: classification report (precision, recall, \(F_1\), support) for each class and macro/weighted averages. Right: confusion matrix (counts). The ensemble aggregates predictions from six tuned models (LR, SVM, RF, Na\"{i}ve Bayes, MLP, XGBoost) with an MLP tie–break, yielding slightly higher overall accuracy than the single MLP while preserving similar class~0 performance.}
  \label{fig:mv_ensemble}
\end{figure}

As shown in Fig.~\ref{fig:mv_ensemble}, the majority-vote ensemble attains an overall accuracy of \textbf{0.857} on the held-out test set. Class-wise performance is strong for class~1 (precision \textbf{0.872}, recall \textbf{0.965}, \(F_1=\textbf{0.916}\)) and comparable to the best single models; for class~0, the ensemble yields precision \(0.727\), recall \(0.400\), \(F_1=0.516\). Relative to the optimized MLP (accuracy \(0.85\), class~0 \(F_1=0.53\)), the ensemble offers a \emph{slightly higher overall accuracy} and class~1 \(F_1\) while maintaining a similar class~0 \(F_1\), but lower recall for class 0. Given our clinical emphasis on identifying class~0, we retain the MLP as the primary model and use the ensemble as a robustness check.

\subsubsection{AdaBoost ensemble}
We also evaluated \emph{AdaBoost}, which builds a stage-wise additive model of weak learners by
reweighting training examples according to previous errors. Starting from uniform weights,
each round fits a base classifier (here, decision stumps/trees); misclassified samples receive
larger weights and correctly classified samples smaller weights, producing a new learner that
focuses on the hardest cases. The final prediction is a weighted vote of all stages, with
weights proportional to each learner's accuracy on reweighted data. 

\begin{figure}[t]
  \centering
  \includegraphics[width=0.95\linewidth]{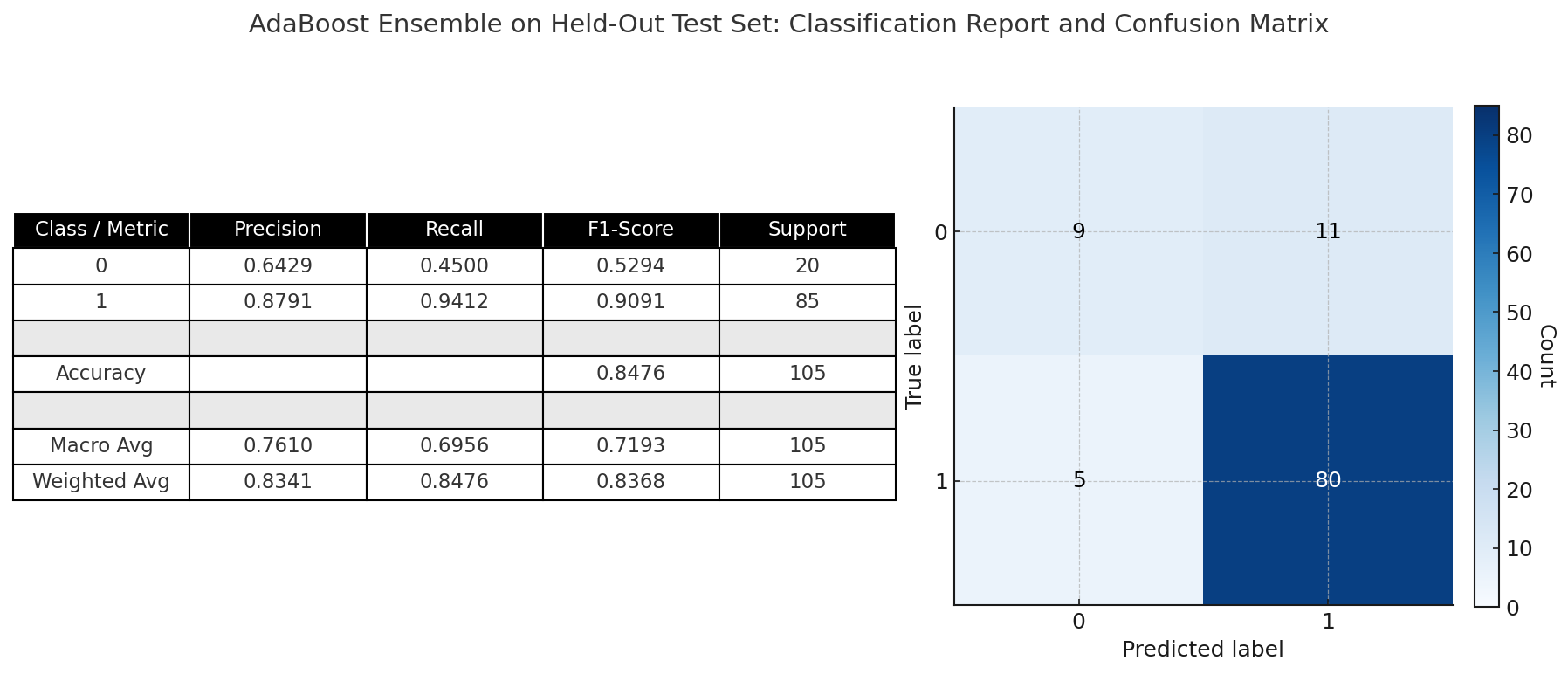}
  \caption{AdaBoost ensemble on the held-out test set. Left: classification report (precision, recall, \(F_1\), support) for each class and macro/weighted averages. Right: confusion matrix (counts) showing balanced overall performance with stronger class~1 detection and moderate class~0 recall.}
  \label{fig:adaboost_results}
\end{figure}

As shown in Fig.~\ref{fig:adaboost_results}, AdaBoost achieved overall accuracy
\(\mathbf{0.8476}\) on the held-out test set. Class-wise performance was strong for class~1 (precision
\(0.8791\), recall \(0.9412\), \(F_{1}=0.9091\)) and moderate for class~0 (precision
\(0.6429\), recall \(0.4500\), \(F_{1}=0.5294\)). Although AdaBoost is competitive
overall, the optimized MLP remains our preferred model because it attains a comparable
accuracy (\(0.85\)) while offering the strongest class~0 \(F_{1}\) among the candidates,
which aligns with our clinical objective of flagging likely non-responders.

\subsubsection{Stacking ensemble}
We constructed a \emph{stacked generalization} (stacking) model that learns how to combine
the out-of-fold predictions from multiple base learners into a final meta-prediction.
The base layer comprised Logistic Regression, SVM, Random Forest, Na\"{i}ve Bayes, MLP,
and XGBoost, each tuned on the training split. For stacking, we generated out-of-fold
predictions from each base model using stratified \(k\)-fold CV on the training data and
used these as features for a meta-learner (logistic regression), which was then trained on
the full out-of-fold matrix and applied to the held-out test set. This design reduces
information leakage and lets the meta-learner exploit complementary error patterns across
models.

\begin{figure}[t]
  \centering
  \includegraphics[width=0.95\linewidth]{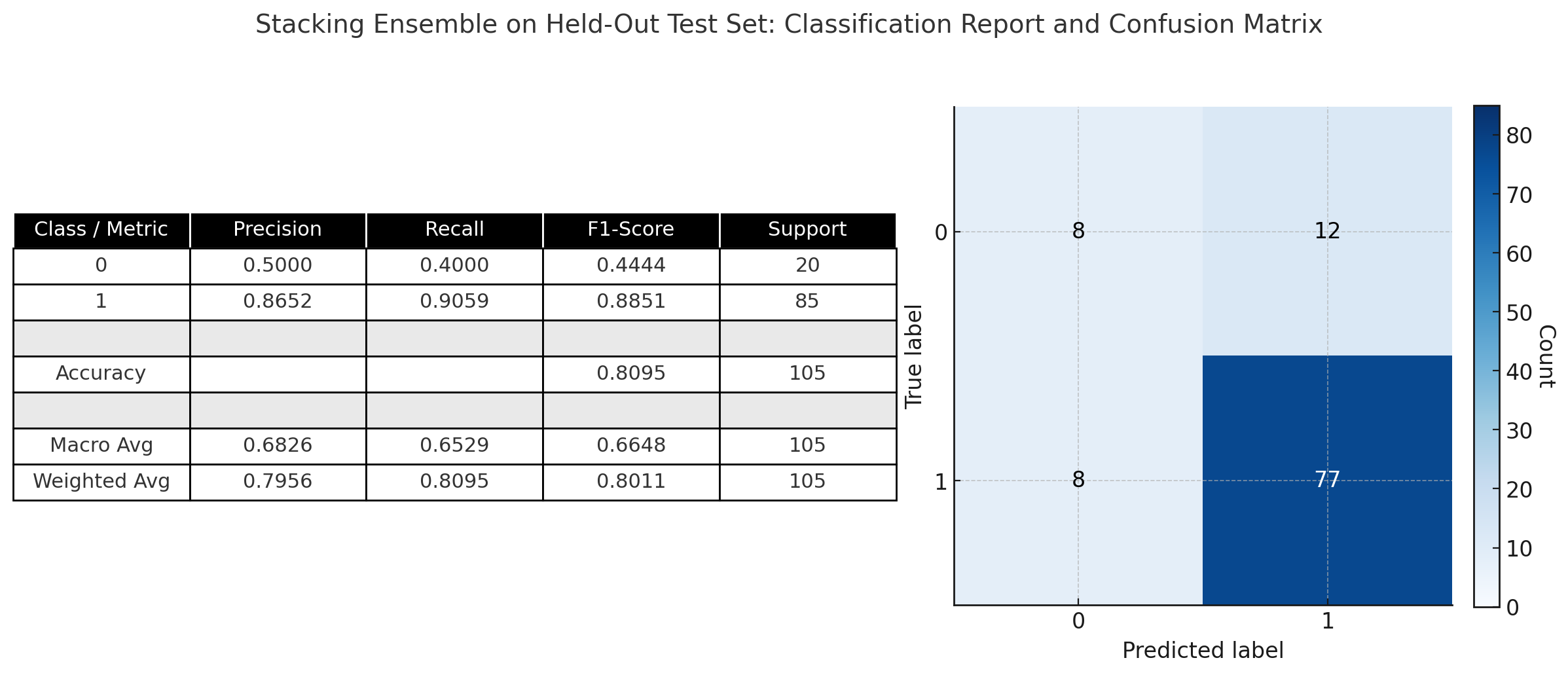}
  \caption{Stacking ensemble on the held-out test set. Left: classification report by class and macro/weighted averages. Right: confusion matrix (counts). The meta-learner combines out-of-fold predictions from six tuned base models (LR, SVM, RF, Na\"{i}ve Bayes, MLP, XGBoost); while overall accuracy is competitive, class~0 detection lags the MLP.}
  \label{fig:stack_results}
\end{figure}

As shown in Fig.~\ref{fig:stack_results}, the stacking ensemble achieved accuracy
\(\mathbf{0.8095}\). Class~1 performance remained strong (precision \(0.8652\),
recall \(0.9059\), \(F_{1}=0.8851\)), while class~0 performance was moderate
(precision \(0.5000\), recall \(0.4000\), \(F_{1}=0.4444\)). Compared with the optimized
MLP (accuracy \(0.85\), class~0 \(F_{1}=0.53\)), stacking did not improve minority-class
detection and thus is not preferred for our clinical objective, though it provides a useful
robustness baseline.

\subsubsection{Can foundation models help? (TabPFN)}
\begin{figure}[t]
  \centering
  \includegraphics[width=0.95\linewidth]{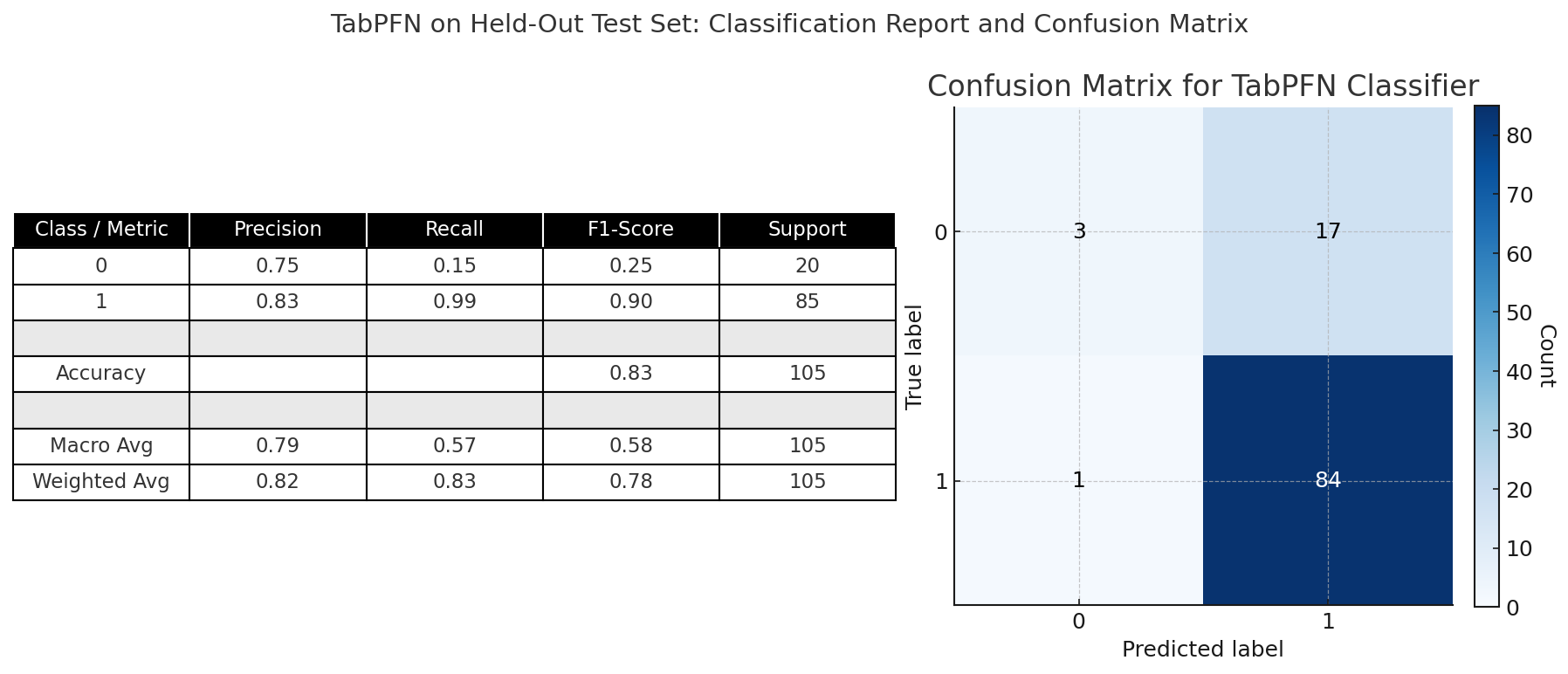}
  \caption{TabPFN on the held-out test set. Left: classification report by class and macro/weighted averages. Right: confusion matrix (counts). While TabPFN delivers excellent class~1 recall (\(0.99\)), class~0 recall (\(0.15\)) lags our optimized MLP, which is prioritized for clinical screening of likely non-responders.}
  \label{fig:tabpfn_results}
\end{figure}

We explored the \emph{TabPFN} foundation model for tabular classification, which uses a
Transformer pre-trained to approximate Bayesian posterior predictive inference on synthetic
tasks and then performs \emph{inference-only} fine-tuning at test time via forward passes
(no gradient updates) \cite{tabpfn}. On our held-out test set, TabPFN achieved accuracy
\(\mathbf{0.83}\) (Fig.~\ref{fig:tabpfn_results}).
Class~1 performance remained high (precision \(0.83\), recall \(0.99\), \(F_{1}=0.90\)),
but class~0 recall was low (\(0.15\); \(F_{1}=0.25\)). Given our clinical objective to
\emph{identify likely non-responders} (class~0), the optimized MLP (class~0 \(F_{1}=0.53\))
remains preferable despite TabPFN’s strong overall discrimination for class~1.

\subsubsection{Benchmarking against human experts and a generative AI baseline}
To validate clinical utility and assess reliability, we benchmarked our model against
\emph{human experts} and a \emph{generative AI} baseline. From the held-out test set we
constructed an evaluation subset of \(n=30\) anonymized CRS cases. We first obtained
class probabilities from the optimized MLP,
\(\hat{p}_i = \Pr(y_i{=}1 \mid \mathbf{x}_i)\), and quantified prediction uncertainty as
\(|\hat{p}_i - 0.5|\). We then stratified by uncertainty to obtain a balanced difficulty
mix: the 10 \emph{most uncertain} cases (smallest \(|\hat{p}_i - 0.5|\)), 10 \emph{most certain}
cases (largest \(|\hat{p}_i - 0.5|\)), and 10
\emph{moderately certain} cases (middle of the remaining list). The final set preserved the approximate class ratio
of the test split and was used for blinded comparison.

\paragraph{Human and LLM baselines}
Six board-certified Otolaryngology physicians who practice as subspecialty Rhinologists independently labeled each case
as \emph{Surgery expected to result in desired improvement in QoL (1)} vs.\ \emph{Surgery will not result in desired improvement in QoL (0)} using the same tabular summary provided
to the model. Surgeons were selected based on participation in the original outcomes study to confirm domain knowledge on CRS surgical outcomes and to ensure familiarity with the clinical data fields. 
To facilitate expert labeling at scale, we deployed a secure web application for case review and adjudication (\url{http://149.166.246.230:4000/patientForms/}). The tool supports authenticated access, session persistence (save-and-resume), and full edit history so clinicians can revise prior responses. A custom graphical user interface (GUI) was developed to collect outcome predictions from these six experienced Rhinologists with expertise in CRS outcomes, all of whom were familiar with the study design and data format. Using the GUI, each expert reviewed detailed subject-level metadata and classified the expected treatment outcome as either “surgical success” or “surgical failure.” Experts also rated their confidence on a five-point Likert scale ranging from “very confident” to “not at all confident.”
 Cases were presented in a standardized tabular format with embedded guidance, and submissions were time-stamped and stored server-side to enable auditability and inter-rater analyses.
In parallel, we evaluated a large language model (ChatGPT, GPT-5
Thinking) using prompt-engineering to elicit a binary decision from the same structured
inputs.\footnote{We did not fine-tune the LLM; prompts asked for a deterministic decision
and brief rationale.}

\begin{figure}[t]
  \centering
  \includegraphics[width=\linewidth]{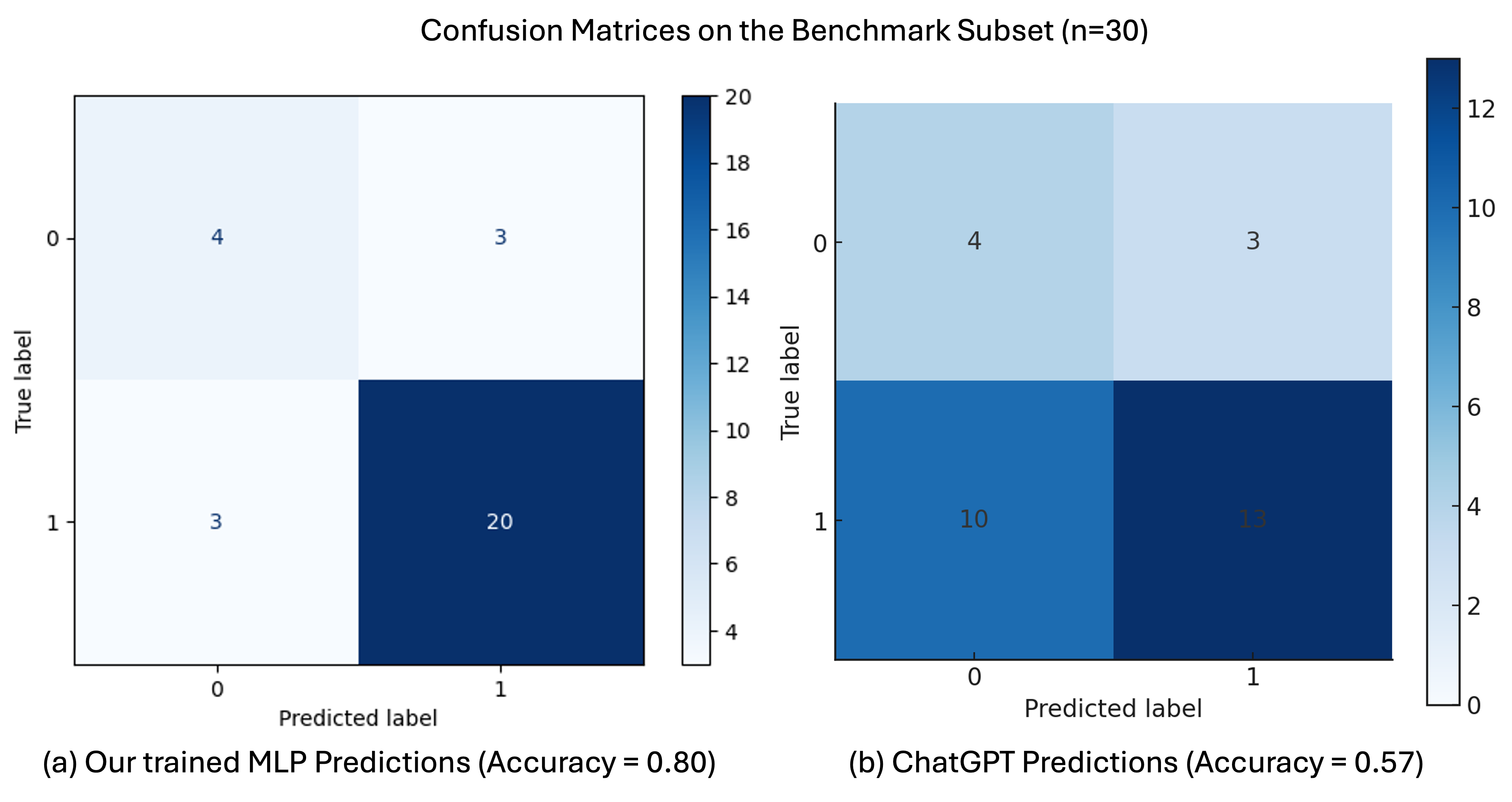}
  \caption{Human-benchmark subset (\(n=30\)). Left (a): MLP prediction confusion matrix
  (\(\mathrm{acc}=0.80\). Right (b): ChatGPT prediction confusion matrix
  (\(\mathrm{acc}=0.57\). The MLP shows markedly higher recall
  for class~1 while maintaining comparable recall for class~0, yielding superior overall
  discrimination on this stratified sample.}
  \label{fig:human_benchmark_cm}
\end{figure}

\paragraph{Results}
Figure~\ref{fig:human_benchmark_cm} shows confusion matrices for our MLP and ChatGPT on
the 30 cases. The MLP achieved
(\(\text{accuracy}=80\%\)), with class-wise F1
metrics of \(\mathrm{F1}_{0}=0.57\), \(\mathrm{F1}_{1}=0.87\), while ChatGPT obtained
\(\text{accuracy}=57\%\),
\(\mathrm{F1}_{0}=0.38\), \(\mathrm{F1}_{1}=0.67\). Notably, both methods recovered
a similar class~0 recall (0.57), but ChatGPT exhibited substantially lower class~1 recall
(0.57 vs.\ 0.87), producing many false negatives (under-calling surgery) in this cohort.

Table~\ref{tab:human_benchmark_split} summarizes accuracies for the expert predictions, our trained MLP and the ChatGPT model. The MLP outperforms the ChatGPT model and performs on par or better than the expert physicians on this subset (the average physician accuracy being 75.6\%), highlighting the model’s potential as a decision-support
tool. Importantly, the uncertainty-stratified sampling increases face validity of the
comparison by ensuring a mix of easy, ambiguous, and hard cases rather than only
``cherry-picked'' examples. 

In addition to reporting individual expert performance, we also formed a \emph{panel-of-experts} prediction for each case by aggregating the six independent expert labels via majority vote. In the event of an exact tie (3 vs.\ 3), we used the experts' self-reported confidence as a tie-breaker by selecting the class with the larger summed confidence across experts voting for that class. This aggregation yields a single consensus decision per case and approximates how multidisciplinary input might be combined in practice.
Table~\ref{tab:expert_panel_metrics} summarizes per-expert metrics and the pooled panel performance on the 30-case benchmarking subset described above. The full six-expert panel achieved 76.7\% accuracy (23/30) with two ties resolved by confidence weighting.
As a sensitivity analysis, we also evaluated a reduced panel consisting of the two highest-accuracy experts on this subset. We again resolved disagreements using the confidence-weighted tie-break rule described above. The resulting top-2 panel performance is reported in Table~\ref{tab:expert_panel_metrics}. Notably, panel aggregation did not improve accuracy relative to the best individual experts, likely because errors were correlated across raters with higher mispredictions, so majority voting gets dominated by shared misinterpretations rather than averaging out independent noise.

\begin{table}[t]
  \centering
  \caption{Thirty-case human benchmarking subset: \textbf{accuracy} by rater. 
  Row~1 reports Doctors~1--4; Row~2 reports Doctors~5--6, the in-house MLP, and ChatGPT.}
  \label{tab:human_benchmark_split}

  %---- Row 1: Doctors 1-4 ----
  \begin{minipage}{\linewidth}
    \centering
    \begin{tabular}{lcccc}
      \toprule
      & \textbf{Doctor 1} & \textbf{Doctor 2} & \textbf{Doctor 3} & \textbf{Doctor 4} \\
      \midrule
      \textbf{Accuracy (\%)} & 76.67 & 66.70 & 70.00 &  83.30  \\
      \bottomrule
    \end{tabular}
  \end{minipage}

  \vspace{6pt}

  %---- Row 2: Doctors 5-6 + models ----
  \begin{minipage}{\linewidth}
    \centering
    \begin{tabular}{lcccc}
      \toprule
      & \textbf{Doctor 5} & \textbf{Doctor 6} & \textbf{MLP (ours)} & \textbf{ChatGPT} \\
      \midrule
      \textbf{Accuracy (\%)} & 80.00 & 76.67 & 80.00 & 56.67 \\
      \bottomrule
    \end{tabular}
  \end{minipage}
\end{table}

\begin{table}[t]
\centering
\caption{Expert benchmarking on the 30-case subset: per-expert performance and panel aggregation. $P_0$ and $P_1$ represent precision for class 0 and 1, respectively, and similarly $R_0$ and $R_1$ represent recall for class 0 and 1, respectively. ``Panel of experts'' uses majority vote; ties are resolved using confidence-weighted voting.}
\label{tab:expert_panel_metrics}
\small
\begin{tabular}{lccccc}
\toprule
\textbf{Rater} & \textbf{Acc} & $\mathbf{P_0}$ & $\mathbf{R_0}$ & $\mathbf{P_1}$ & $\mathbf{R_1}$ \\
\midrule
Doctor 1   & 0.767 & 0.500 & 0.286 & 0.808 & 0.913 \\
Doctor 2 & 0.667 & 0.286 & 0.286 & 0.783 & 0.783 \\
Doctor 3   & 0.700 & 0.333 & 0.286 & 0.792 & 0.826 \\
Doctor 4      & 0.833 & 0.750 & 0.429 & 0.846 & 0.957 \\
Doctor 5    & 0.800 & 0.667 & 0.286 & 0.815 & 0.957 \\
Doctor 6  & 0.767 & 0.500 & 0.143 & 0.786 & 0.957 \\
\midrule
Panel of experts (6)     & 0.767 & 0.500 & 0.143 & 0.786 & 0.957 \\
Top-2 panel (Doctors 4 and 5)  & 0.800 & 0.667 & 0.286 & 0.815 & 0.957 \\
\bottomrule
\end{tabular}
\end{table}

\begin{figure}[t]
\centering
\includegraphics[width=0.72\linewidth]{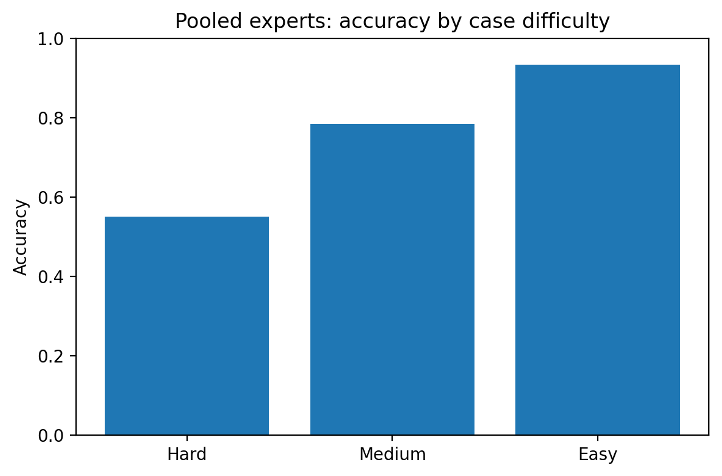}
\caption{Pooled experts: accuracy by case difficulty tier (Hard/Medium/Easy) on the 30-case benchmark subset.}
\label{fig:expert_difficulty_bar}
\end{figure}

\begin{figure}[t]
\centering
\includegraphics[width=0.90\linewidth]{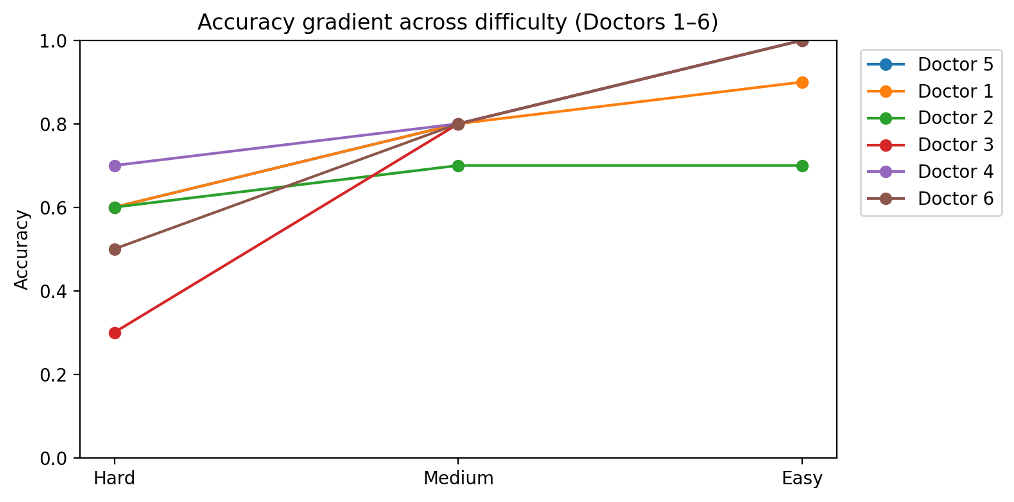}
\caption{Accuracy gradient across difficulty tiers for each expert and the pooled mean. Most raters show a monotonic Hard$\rightarrow$Medium$\rightarrow$Easy increase.}
\label{fig:expert_difficulty_lines}
\end{figure}

\paragraph{Performance gradient across case difficulty (Hard $\rightarrow$ Medium $\rightarrow$ Easy)}
Because the 30-case benchmark set was constructed by stratifying cases according to the MLP's prediction uncertainty, it naturally supports an analysis of whether both human and model performance varies across difficulty tiers.  
When pooling expert judgments within each tier, accuracy increased monotonically from Hard to Easy (average accuracy - Hard: 0.55; Medium: 0.783; Easy: 0.933), indicating a clear difficulty gradient (Fig.~\ref{fig:expert_difficulty_bar}). Figure~\ref{fig:expert_difficulty_lines} further shows that most individual experts exhibit the same Hard$\rightarrow$Medium$\rightarrow$Easy pattern, with the pooled mean tracking this monotonic rise.

The optimized MLP exhibited a similarly monotonic gradient across the same tiers, achieving accuracies of 0.60 (Hard), 0.80 (Medium), and 1.00 (Easy). Together, these results suggest that both clinician and model errors concentrate in the most ambiguous (Hard) cases, while performance approaches ceiling on highly certain (Easy) cases.

In terms of comparison with the generative AI model, across the uncertainty spectrum, our MLP maintained balanced performance and substantially
higher sensitivity for class~1, while ChatGPT tended to under-call surgery (false negatives).
With comparable performance to physicians in terms of prediction accuracy, the MLP’s \(+23.3\) percentage
point advantage over ChatGPT (80.0\% vs.\ 56.7\%) and its strong class-wise F1 scores further
support its applicability as a complementary decision-support tool to aid clinicians, rather
than a replacement for expert judgment.

\begin{figure}[t]
\centering
\includegraphics[width=\columnwidth]{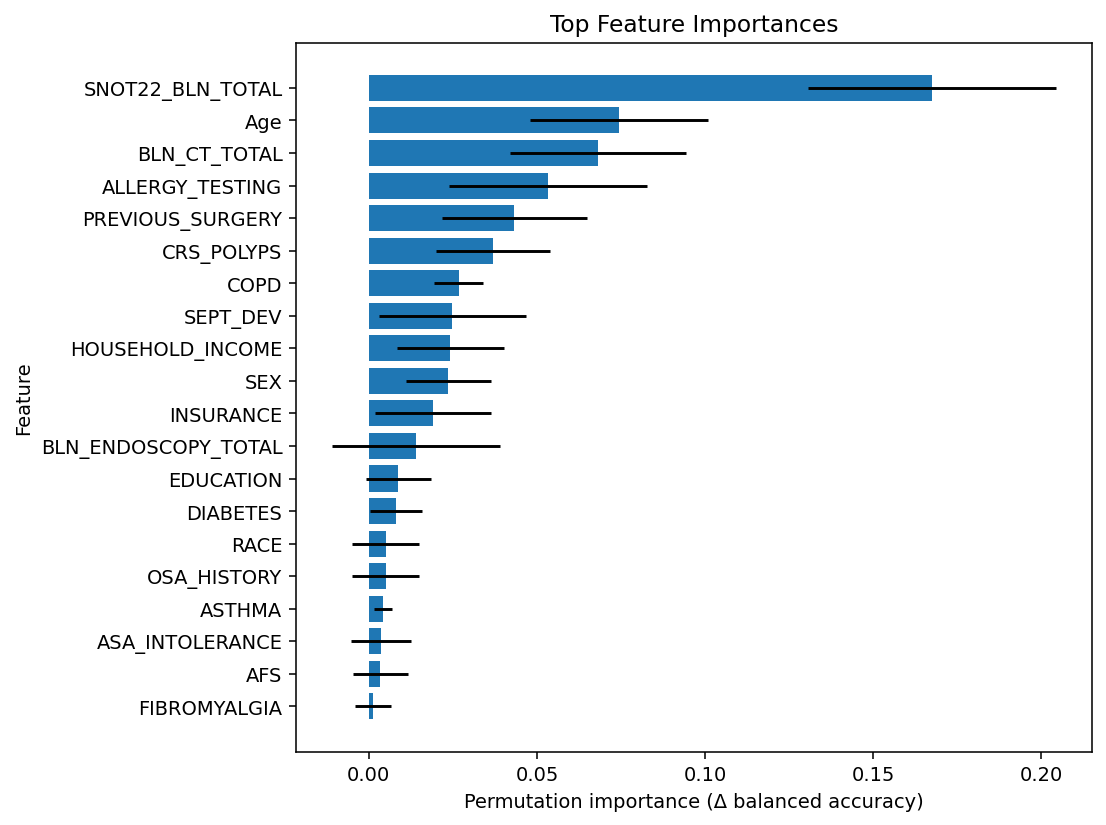}
\caption{Permutation feature importance for the MLP classifier on the held-out test set. Bars show the mean decrease in balanced accuracy when each feature is randomly permuted (larger values indicate greater importance); black ticks denote the standard deviation across permutation repeats. The model is most sensitive to \texttt{SNOT22\_BLN\_TOTAL}, \texttt{Age}, and \texttt{BLN\_CT\_TOTAL}, followed by \texttt{ALLERGY\_TESTING}, \texttt{PREVIOUS\_SURGERY}, and \texttt{CRS\_POLYPS}.}
\label{fig:mlp_feature_importance}
\end{figure}

\subsubsection{Feature importance analysis}
Understanding which variables drive model predictions is essential for clinical adoption and
for generating testable hypotheses. We therefore quantified \emph{global} feature importance
for the optimized MLP using \textbf{permutation importance} on the held-out test set, scored
by the drop in balanced accuracy (\(\Delta\)~BA) after randomly permuting each feature while
holding all others fixed. To reduce Monte Carlo noise, we averaged importance over multiple
permutations per feature and report the mean \(\pm\)~standard deviation as error bars
(Fig.~\ref{fig:mlp_feature_importance}). This procedure probes the model’s \emph{reliance}
on each input without assuming linearity and is invariant to feature scaling.

\paragraph{Key findings}
The baseline SNOT-22 total score (\texttt{SNOT22\_BLN\_TOTAL}) was by far the most influential
predictor, producing the largest decrease in balanced accuracy when permuted. This aligns with
clinical intuition: preoperative symptom burden is strongly associated with postoperative
patient-reported outcomes and thus with the model’s recommendation signal. Age and imaging disease
burden (baseline CT score, \texttt{BLN\_CT\_TOTAL}) were the next most important, consistent with known
associations between disease severity, remodeling, and treatment response. Positive allergy testing and prior surgery also ranked highly, suggesting that atopy and surgical history
modulate expected benefit. Additional contributors included presence of nasal polyps
(\texttt{CRS\_POLYPS}), COPD, and septal deviation (\texttt{SEPT\_DEV}), all biologically
plausible correlates of disease complexity and airflow/ventilation. Socioeconomic proxies
(\texttt{HOUSEHOLD\_INCOME}, \texttt{INSURANCE}) and demographics (\texttt{SEX}, \texttt{RACE})
showed modest but nonzero importance, indicating potential potential social determinants of health patterns; these variables should be monitored for causality rather than
interpreted causally. Lower-ranked comorbidities (e.g., diabetes, OSA, asthma) contributed
incrementally, consistent with their secondary roles in CRS symptomatology.

\paragraph{Clinical plausibility and caveats}
Overall, the ranking mirrors domain expectations: symptoms \(\rightarrow\) anatomic/imaging
severity \(\rightarrow\) atopy/surgical history \(\rightarrow\) comorbidity/socioeconomic context.
Because permutation importance reflects the model’s \emph{predictive dependence} and not causal
effects, correlated features can share or mask importance, and interactions may distribute
influence across variables. Accordingly, we treat these results as a global sanity check and
pair them with local explanations (e.g., SHAP) next.
This combination improves interpretability and supports responsible deployment in
decision-support workflows.

\subsubsection{Global and local interpretability via SHAP}
\begin{figure}[t]
\centering
\includegraphics[width=\columnwidth]{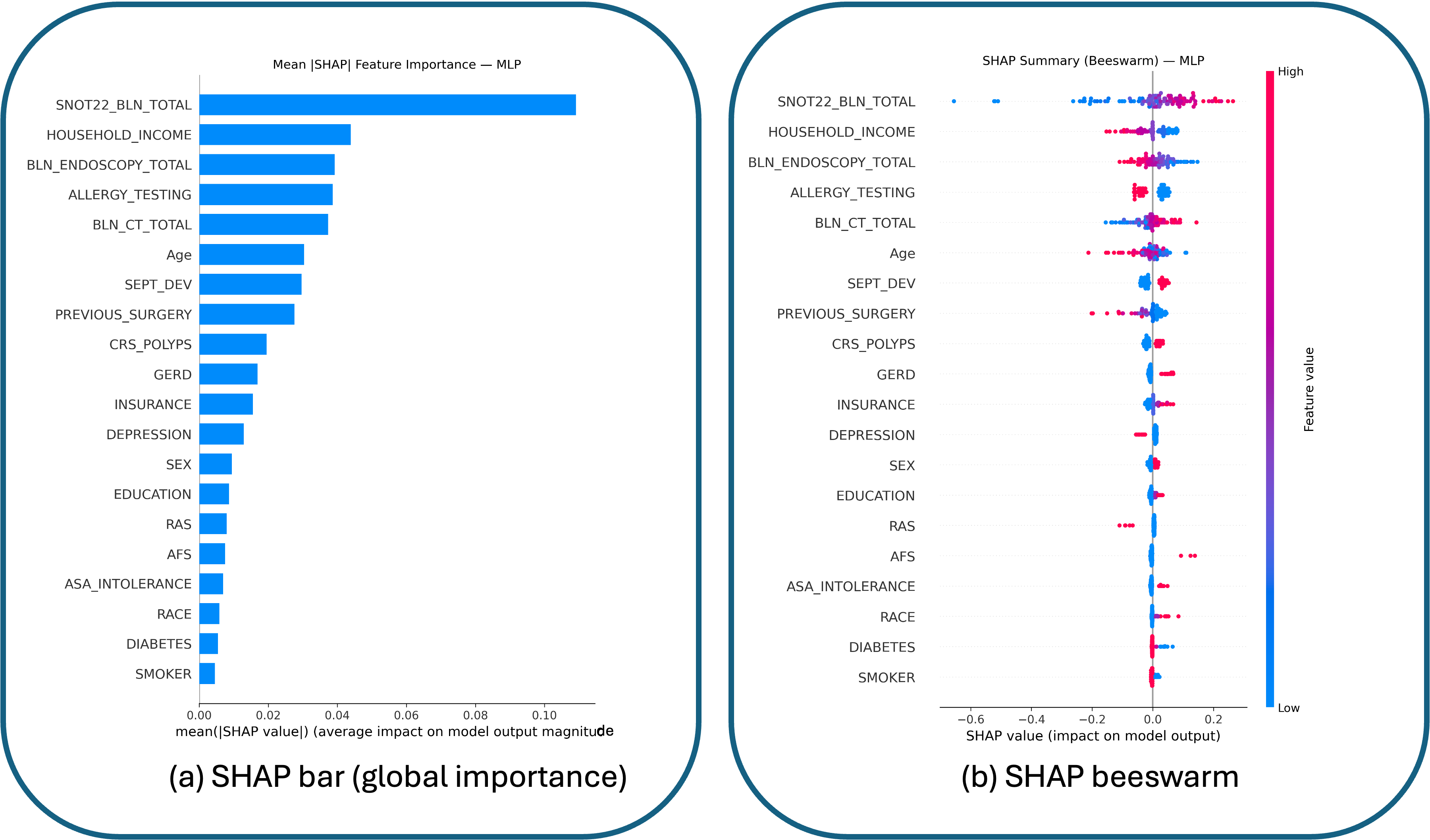}
\caption{(a) SHAP bar (global importance): Mean absolute SHAP values on the test set rank features by their average contribution to the predicted probability of achieving the desired outcome; higher bars indicate greater global influence. (b) SHAP beeswarm: Each point represents a patient; the horizontal axis shows the SHAP value (feature effect on the log-odds/probability), and color encodes the feature value from low$\to$high. Distributions reveal both effect size and heterogeneity.}
\label{fig:shap_vals}
\end{figure}

To complement permutation importance and provide \emph{directional}, patient-level
explanations, we computed Shapley additive explanations (SHAP) for the optimized MLP.
SHAP builds on the Shapley value from cooperative game theory~\cite{Shapley1953}
and provides consistent, locally faithful attributions for ML models~\cite{LundbergLee2017}.
 We used the standard Kernel/Deep SHAP interface on the held-out test set and
summarize results in Fig.~\ref{fig:shap_vals}.

\paragraph{Global ranking and directionality}
The mean-$|\mathrm{SHAP}|$ bar plot (left) broadly corroborates the permutation ranking:
the SNOT-22 baseline score dominates, followed by household income, 
baseline endoscopy score, positive allergy testing, baseline CT score, and
age, with anatomic factors (e.g., septal deviation), history (e.g.,
previous surgery), and polyps contributing next.
The beeswarm plot (right) adds \emph{sign}: high preoperative symptom burden
(shown with SNOT-22 baseline score, red points) concentrates on the positive SHAP side, pushing
the model toward \(\hat{y}{=}1\) (surgery), whereas low values (blue) push toward
\(\hat{y}{=}0\). A similar trend is visible for objective disease measures
(baseline endoscopy score, baseline CT score) and atopy
(positive allergy testing); higher values generally increase the predicted probability
of surgery, consistent with domain expectations that greater symptom and anatomic burden
favor operative management.

\paragraph{Heterogeneity and fairness considerations}
Socioeconomic variables (like household income, insurance) exhibit
substantial spread in SHAP values, indicating heterogeneous effects across patients. These social factors are important to surgical outcomes as they determine care access, medicine availability, diet, and air quality. 
While these features improve discrimination, they can also encode access or practice
patterns rather than biology; we therefore interpret them cautiously, monitor their
contributions, and report fairness diagnostics separately. Demographics (sex,
race) and several comorbidities (e.g., diabetes, smoker)
show smaller average contributions, aligning with their lower global importances.

\paragraph{Clinical plausibility and use}
Together, SHAP and permutation analyses produce a coherent narrative: symptoms and
endoscopic/CT severity are primary drivers, modulated by atopy and surgical history,
with comorbidities and socioeconomic context providing secondary signal. SHAP’s local
attributions are particularly useful for case review: clinicians can see \emph{which}
factors pushed an individual recommendation toward surgery vs.\ no surgery, supporting
transparent, auditable decision support. These also aid in interpretability in a real-time clinical decision support application in community health settings where there is limited time and perhaps the doctor and the patient are not particularly familiar with data science techniques.

\section{Discussion}
This work demonstrates that routinely available preoperative data can support clinically useful individualized prediction of endoscopic sinus surgery (ESS) outcomes in Chronic Rhinosinusitis (CRS). Among multiple supervised learners, an optimized multilayer perceptron (MLP) provided the best balance of overall accuracy (\(\approx 0.85\) on the held-out test set) and minority-class detection (class~0 F1 \(\approx 0.53\)), outperforming classic baselines (logistic regression, SVM, naive Bayes, random forest) and more sophisticated alternatives (XGBoost, stacking/boosting ensembles). Gains from ensembling were modest, consistent with a regime in which model variance is already low relative to the noise and sample size constraints of the problem. Human benchmarking on a stratified 30-case subset further showed that the MLP exceeded a strong generative-AI baseline and performed comparably to expert physicians, indicating that signal present in standard clinical variables is sufficient to aid decision support. Additionally, the human expert feedback provides valuable clinician intuition and enables comparing the factors deemed as salient by the ML model and the expert.

\paragraph{Interpretability and clinical plausibility}
Interpretability was a desired property of our modeling. Global permutation importance and SHAP analyses converged on a clinically coherent hierarchy: disease severity measures including patient reported SNOT-22, objective clinician scores for CT and endoscopy exams, and atopy/surgical history were the principal drivers of the model’s recommendations, with comorbidities and socioeconomic context providing secondary signal. There can be a complex and individualized interplay among these many factors, the exploration of which could be an interesting future work. 
%A web application exposes these dependencies through interactive response surfaces and patient-level explanations, enabling ``what–if'' analyses in two conditioning modes (global cross-section vs.\ on-manifold interpolation). This tooling makes the model’s behavior auditable and facilitates shared decision making.

\paragraph{Class imbalance and error asymmetry}
Across all algorithms, performance on class~0 (patients unlikely to achieve the desired outcome) trailed class~1. This asymmetry is expected: class~0 is less frequent and potentially more heterogeneous. In many medical outcome settings, minority classes represent non-response or adverse events that are under-sampled, noisier, and driven by unmeasured factors. Our results reflect this: while class~1 F1 \(\approx 0.91\), class~0 F1 \(\approx 0.53\) for the MLP. We tried to mitigated the imbalance through class weighting, but a fundamental data limitation remains. Future work will explicitly address imbalance with %cost-sensitive learning and focal losses; 
re-sampling (oversampling of the minority class) and synthetic data generation (e.g., SMOTE/ADASYN, mixup); and model-based augmentation (e.g., VAEs/GANs) constrained to clinically plausible regions of feature space. Active learning with expert feedback focused on uncertain/rare cases could further enrich the minority class efficiently.

\paragraph{Limits of the current study}
Several considerations exist, of course. First, the dataset is moderate in size. Although we used conservative preprocessing and assessed robustness, sample size likely caps achievable accuracy and minority recall. Second, our endpoint is binarized at six months. Some non-responders may benefit later, and continuous or longitudinal outcomes (e.g., \(\Delta\)SNOT-22 trajectories) could be more informative, although our prior study shows this is limited \cite{deconde2015longitudinal}. Third, socioeconomic variables improved prediction but can encode access or practice patterns. As such, they should be treated as potential confounders and subgroup performance should be monitored for fairness, which we plan to do in the future. Fourth, while the human benchmarking is encouraging, it was conducted on a small stratified subset of samples against legitimate experts in the field. We do not know how the average physicians in the community will compare, who have different experiences and may not keep up with the nuances in the literature or dive deeply into such conversations. In terms of the generative AI comparison, we experimented with a single LLM configuration. Overall, broader reader studies and varied LLM baselines are warranted. Finally, external generalizability remains to be proven; site effects, referral patterns, surgical technique and perioperative care, medication adherence, patient psychology, priming effects for survey responses are all considerations that may shift feature distributions.

\paragraph{Clinical relevance}
Even with these constraints, \(\approx 85\%\) accuracy with transparent explanations is meaningful for preoperative counseling. Errors are not equivalent clinically: false negatives (predicting non-benefit when benefit is likely) may delay effective surgery, while false positives may expose a patient to operative risk with limited expected gain. Our framework supports threshold tuning and individualized decision curves, enabling clinicians to select operating points aligned with patient preferences and risk tolerance. The interactive visualizations help communicate why the recommendation was made and how it might change under alternative assumptions.

\paragraph{Future directions}
We outline several priorities to translate these findings:
\begin{enumerate}
  \item \textbf{Data growth and harmonization:} expand to broader cohorts; standardize variable definitions; leverage federated learning for privacy-preserving training. Also, include other surgery specific and patient psychological factors, other outcomes of relevance for decision making.
  \item \textbf{Learning under imbalance:} integrate calibrated re-weighting, adopt techniques like SMOTE/ADASYN/mixup, and generative augmentation with clinical plausibility checks; evaluate minority-aware metrics prospectively.
  \item \textbf{Temporal/external validation:} perform temporal splits, external site validation, and pre-registered prospective studies; assess domain shift and transportability.
  \item \textbf{Better targets and calibration:} model continuous and time-to-event endpoints; report Brier score/ECE; apply temperature/Platt calibration; use decision-curve analysis to quantify net benefit across thresholds.
  \item \textbf{Uncertainty and reliability:} add Bayesian/post-hoc uncertainty quantification, conformal prediction, and reject-option policies to flag cases for multidisciplinary review.
  \item \textbf{Individualized treatment benefit:} move beyond outcome prediction to estimate conditional treatment effects (uplift models, causal forests) to quantify patient-specific expected gain from surgery versus medical therapy.
  \item \textbf{Human--AI collaboration:} expand reader studies, measure assistance effects (with/without model), and optimize the UI for rapid, explainable triage; incorporate clinician feedback loops for continual learning.
  \item \textbf{Foundation/self-supervised models:} our TabPFN exploration showed strong class~1 discrimination but limited minority recall; future work will test tabular foundation models and hybrid pipelines (self-supervised feature learning + calibrated discriminators) under imbalance constraints.
\end{enumerate}

\section{Conclusion}
We present an interpretable, data-efficient pipeline for predicting CRS surgical outcomes. An optimized MLP achieved strong discrimination on a held-out cohort and surpassed both a generative-AI baseline and expected human performance on a stratified reader set, while providing transparent global and local explanations. The principal limitation is minority-class performance driven by class imbalance and sample size, ubiquitous challenges in medical outcome prediction. Addressing these with targeted data growth, imbalance-aware learning, rigorous external validation, and careful calibration will be key to translating such models from retrospective studies to prospective, decision-support tools that improve patient counseling and, ultimately, outcomes.

\section*{ACKNOWLEDGMENT}

This work was supported by a grant from the National Institute of Allergy and Infectious Diseases (NIAID) of the National Institutes of Health under grant number R01AI175631 (VRR). Contents are the authors’ sole responsibility and do not necessarily represent official NIH views.

%% The Appendices part is started with the command \appendix;
%% appendix sections are then done as normal sections
\appendix
\section{Tripod AI Check}
\label{app1}
\textcolor{red}{I checked mostly, we may need to confirm 1 point: fairness.}

%% If you have bib database file and want bibtex to generate the
%% bibitems, please use
%%
%%  \bibliographystyle{elsarticle-num} 
%%  \bibliography{<your bibdatabase>}

%% else use the following coding to input the bibitems directly in the
%% TeX file.

%% Refer following link for more details about bibliography and citations.
%% https://en.wikibooks.org/wiki/LaTeX/Bibliography_Management

%% References with bibTeX database:
\bibliographystyle{ieeetr}
\bibliography{sample}

\end{document}